\pdfoutput=1
\documentclass[11pt]{article}
\usepackage[final]{acl}

\usepackage{times}
\usepackage{latexsym}
\usepackage[T1]{fontenc}
\usepackage[utf8]{inputenc}
\usepackage{microtype}
\usepackage{inconsolata}
\usepackage{graphicx}
\usepackage{longtable}
\usepackage{float}
\usepackage[most]{tcolorbox}
\usepackage{subcaption}
\usepackage{enumitem}
\usepackage{hyperref}
\usepackage{booktabs}
\usepackage{multirow}
\usepackage{makecell}
\usepackage{array}
\usepackage[english]{babel}
\usepackage{inconsolata}
\usepackage{amsmath}
\usepackage{amssymb}
\usepackage{pifont}
\usepackage{xcolor}
\usepackage[normalem]{ulem}

\definecolor{mainbg}{HTML}{e7f5ff}
\definecolor{titlebg}{HTML}{a5d8ff}
\definecolor{fontcolor}{HTML}{1971c2}

\title{Crossroads of Continents: Automated Artifact Extraction for Cultural Adaptation with Large Multimodal Models}

\author{
Anjishnu Mukherjee \quad
Ziwei Zhu\textsuperscript \quad 
Antonios Anastasopoulos \\
Department of Computer Science, George Mason University \\
\texttt{\{amukher6,zzhu20,antonis\}@gmu.edu}
}

\newcommand{\dataset}{\textsc{Dalle~Street}}
\def \gpt4v{GPT-4V}
\def \dalle{DALL-E~3}
\def \llava{LLaVA}
\def \dino{Grounding DINO}
\def \sd{Stable Diffusion}

\def \clipscore{CLIPScore}
\def \geoguessr{GeoGuessr}
\def \marvl{\textsc{MaRVL}}
\def \dollarstreet{\textsc{Dollar~Street}}
\def \cultureadapt{\textsc{CultureAdapt}}

\begin{document}
\maketitle

\begin{abstract}
We present a comprehensive three-phase study to examine ($1$) the cultural understanding of Large Multimodal Models (LMMs) by introducing \dataset{}, a large-scale dataset generated by \dalle{} and validated by humans, containing $9{,}935$ images of $67$ countries and $10$ concept classes; ($2$) the underlying implicit and potentially stereotypical cultural associations with a cultural artifact extraction task; and ($3$) an approach to adapt cultural representation in an image based on extracted associations using a modular pipeline, \cultureadapt{}. We find disparities in cultural understanding at geographic sub-region levels with both open-source (\llava{}) and closed-source (\gpt4v{}) models on \dataset{} and other existing benchmarks, which we try to understand using over $18{,}000$ artifacts that we identify in association to different countries. Our findings reveal a nuanced picture of the cultural competence of LMMs, highlighting the need to develop culture-aware systems.\footnote{Dataset and code are available: \url{https://github.com/iamshnoo/crossroads}}
\end{abstract}

\section{Introduction}
\label{sec:intro}

\begin{figure}[t]
    \centering
    \includegraphics[width=0.48\textwidth]{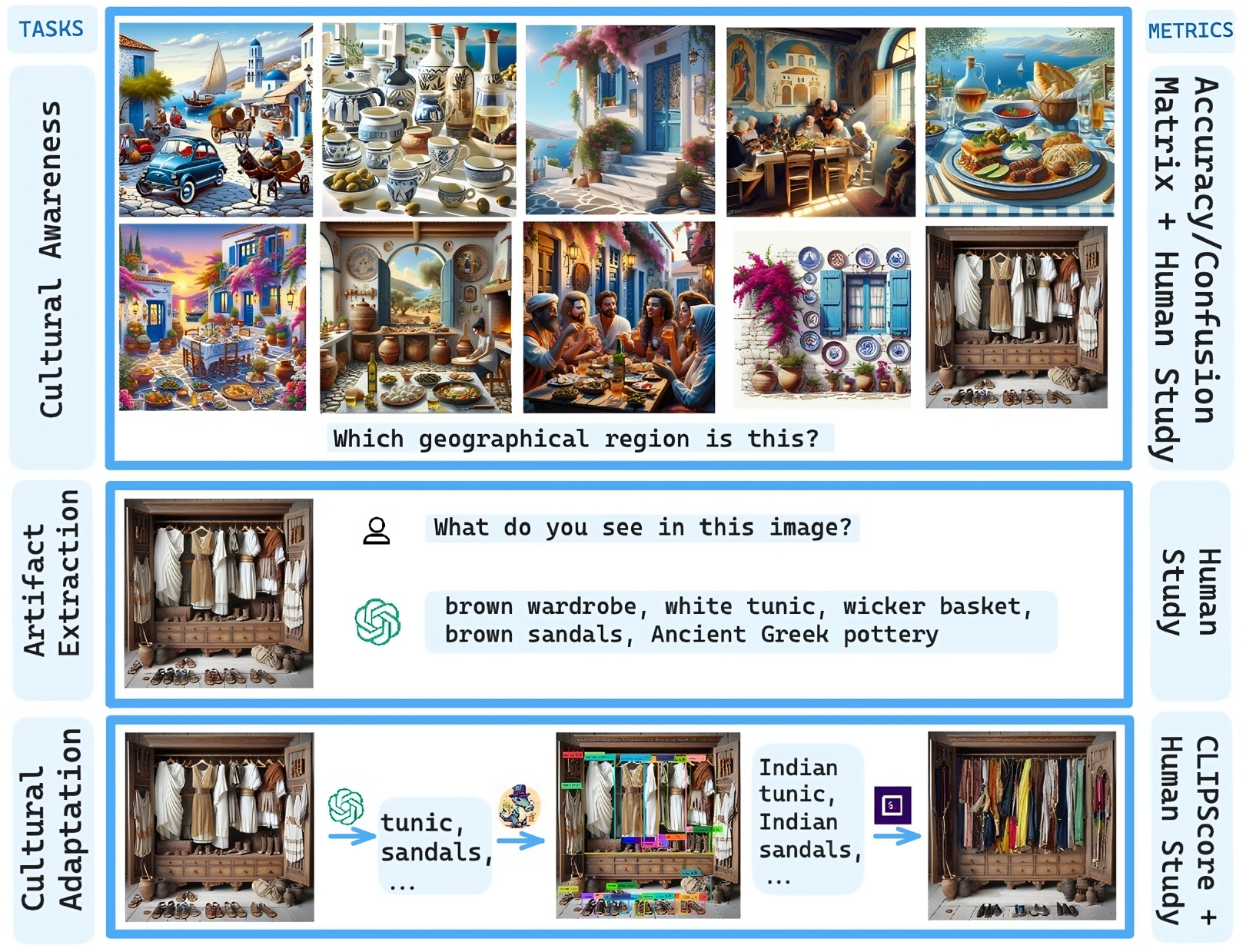}
    \caption{We introduce a large-scale dataset for measuring cultural awareness, an artifact extraction task for implicit cultural associations, and a modular pipeline for culturally adapting images with fine-grained edits.}
    \label{fig:fig1}
    \vspace{-1em}
\end{figure}

Culture is hard to define and has always been so. \citet{kroeber-1952-culture} explored how the word has evolved to gain different meanings in different contexts. Recent efforts in natural language processing research have seen growing interest in understanding how culture influences language models and human behavior, including language, art, and decision-making \cite{hershcovich-2022-challenges, adilazuarda-2024-towards, liu-2024-culturally, ge-2024-culture-shapes}. As large multimodal models (LMMs) intersect more with human life, the need for them to comprehend and respect cultural nuances is crucial.
Research in this area focuses on model alignment with human values, assessing \textit{cultural awareness}, and exploring \textit{cultural adaptation}; the goal is to modify content that represents one culture or country, often stereotypically, to reflect a different one, to suit audiences from different cultural backgrounds better.

The challenge of assessing the capability of understanding and leveraging cultural knowledge in LMMs is significant. Prior research has primarily investigated LMMs for cultural awareness\footnote{We maintain that LMMs do not inherently possess human values but that their outputs may display cultural knowledge.} by examining their performance on tasks such as region classification from images \cite{basu-2023-inspecting, yin-2023-givl, pouget-2024-no-filter}, image-caption matching \cite{liu-2021-marvl}, and cultural image captioning~\cite{cao-2024-exploring}. However, these tasks do not determine whether LMMs are responding to cultural cues encoded within their training data or merely identifying superficial cultural associations. 

To address these gaps, first, we develop a new large-scale \textbf{dataset} to assess cultural awareness as measured by the ability of LMMs to recognize and differentiate between cultures, with countries as proxies, in a task setting similar to \geoguessr{} \cite{geoguessr-2024}. Next, we introduce a \textbf{task} designed to identify implicit associations between cultures and artifacts \cite{liu-2024-culturally} that LMMs use to distinguish between cultures. Finally, we propose a \textbf{cultural adaptation framework} combining multiple generative models in an end-to-end pipeline to adapt images from one cultural context to another by modifying the underlying implicit associations. Our main contributions are as follows:

\begin{itemize}[nolistsep,noitemsep,leftmargin=*]
    \item \textbf{Dataset:} We introduce \dataset, a collection of $9{,}935$ images generated by \dalle{}, covering $67$ countries and $10$ cultural concept classes, with more images from underrepresented geographic regions compared to datasets like \dollarstreet{} \cite{rojas-2022-dollarstreet}.
    \item \textbf{Benchmark:} We measure how well humans and multimodal large language models (open- and closed-source) can identify countries for images in \dataset{} and two other datasets (\dollarstreet, \marvl) to study disparities in performance at the geographic subregion level for a diverse group of concepts and~countries.
    \item \textbf{Task:} We introduce a task for identifying implicit associations by extracting cultural artifacts from images and filtering them to discover associations that frequently co-occur for each country.
    \item \textbf{Framework:} We propose a modular end-to-end pipeline, \textsc{CultureAdapt} (Figure~\ref{fig:culture-adapt-vs-dalle}), to adapt an image to a target culture by updating identified implicit cultural associations in it using diffusion-based inpainting. We evaluate results by introducing a \clipscore{}-based metric. 
\end{itemize}

\section{Data}
\label{sec:data}

We study around $20$k images from three datasets (data statistics in Table~\ref{tab:statistics-sub-regions}) covering a wide variety of cultural concepts, economic ranges, and data sources: (a) \dataset{}: synthetic, \dalle{} \cite{openai-2024-dalle3} generated; (b) \dollarstreet{}: natural, collected photographs; and (c) \marvl{}: web-scraped under native speaker guidance. All the datasets are available under CC~BY-SA~$4.0$ license.

\paragraph{\dataset{}} We use $10$ concept classes (car, family snapshots, front door, home, kitchen, plate of food, cups/mugs/glasses, social drink, wall decoration, and wardrobe), $19$ geographical regions, and $67$ countries (Section~\ref{sec:data-info}), similar to \dollarstreet{}. We generate $1024\times1024$ images with \dalle{}~\cite{openai-2024-dalle3} in two styles - vivid (hyper-realistic) and natural (realistic), prompting with a template (Figure~\ref{fig:dalle-prompt}) that specifies the concept class and target country. At least $10$ images are sampled per country-concept combination, yielding $9{,}935$ images after filtering out content policy violations from API calls. A qualitative study on a randomly sampled subset of around $300$ generated images and $14$ participants (Table~\ref{tab:human-appropriateness}) shows most annotators agree that the images reflect stereotypical country representations, with less than $1\%$ of images receiving strong disagreement. When unsure, participants tend to neither agree nor disagree about the ``appropriateness'' of an image. Our annotators also mark \textit{visual cues} in these images, including both explicit (e.g., flags) and implicit (e.g., color schemes) cultural artifacts. This feedback motivates our artifact extraction task and its use in our cultural adaptation framework.

\paragraph{\dollarstreet{}~\cite{rojas-2022-dollarstreet}} This is a dataset of photos of objects and scenes collected by professional and volunteer photographers. We filter it for images that do not contain multiple labels, classes that do not cover images for all regions, and classes with subjective naming. Then we refer to our group of annotators from diverse backgrounds to choose the top $10$ categories by the method of collaborative labeling~\cite{chang-2017-collaborative-labeling}, where we simplify our selection of object classes by choosing the ones which all annotators universally agree on as being a relevant dimension for testing cultural awareness. Our data from this source includes $4{,}137$ images from $63$ countries, $19$ geographical regions, across $10$ concept classes.

\paragraph{\marvl{} \cite{liu-2021-marvl}} This is originally a dataset for validation of statements about image pairs curated by native speakers in five languages: Indonesian, Mandarin Chinese, Swahili, Tamil, and Turkish. We assign country and region labels (based on corresponding languages) to get $4{,}914$ images across~$5$ geographical regions.

\section{Cultural Awareness (Task 1)} 
\label{sec:cultural-awareness}
We compare performances of humans, \llava{} and \gpt4v{} on \dataset{}, \dollarstreet{} and \marvl{} in terms of their ability to predict the country given an image. Overall, we find performances vary across sub-regions, but both \llava{} and \gpt4v{} perform better than humans.

\begin{table}[t]
    \small
    \centering
    \begin{subtable}[b]{0.45\textwidth}
        \centering
        \begin{tabular}{lc}
            \toprule
            \textbf{Appropriateness Category} & \textbf{Percentage (\%)} \\
            \midrule
            Agree                    & $40.79$  \\
            Neither Agree nor Disagree & $34.54$  \\
            Strongly Agree            & $19.41$   \\
            Disagree                  & $4.61$   \\
            Strongly Disagree         & $0.66$    \\
            \bottomrule
        \end{tabular}
        \caption{Results from our human study on appropriateness for generated images show that most participants agree or are neutral, with less than $1$\% expressing strong disagreement.}
        \label{tab:human-appropriateness}
    \end{subtable}
    \begin{subtable}[b]{0.45\textwidth}
        \centering
        \begin{tabular}{lc}
            \toprule
            \textbf{Geographical Level} & \textbf{Accuracy (\%)} \\
            \midrule
            Country Level        & $22.16$ \\
            Subregion Level      & $47.63$ \\
            Continent Level      & $77.77$ \\
            Union Accuracy       & $78.03$ \\
            Intersection Accuracy & $21.91$ \\
            \bottomrule
        \end{tabular}
        \caption{Accuracy for the cultural awareness task improves from country to subregion to region level.}
        \label{tab:human-accuracy}
    \end{subtable}
    \caption{(a) Perceived appropriateness of generated images by human participants. (b) Cultural awareness accuracy at different geographical levels for a subset of \dataset{} images.}
    \label{tab:human-study}
    \vspace{-1em}
\end{table}

\subsection{Methods}
Given an input image, we prompt \texttt{\llava-NeXT} \cite{haotian-2023-llava} and \gpt4v{} \texttt{vision-preview} \cite{openai-2023-gpt4v} in a zero-shot generative setting by asking an open-ended question without providing answer choices: \textit{Predict the geographical region represented in the image, as per the United Nations geoscheme}~\cite{unsd-geoscheme-2024}. We use this geoscheme for three reasons: ($1$) models have a higher refusal rate when queried with specific country labels; ($2$) the geoscheme is included in most LLM pre-training data (English Wikipedia); and ($3$) it enables structured parsing of open-ended generations. We focus on countries with stable geographic classifications to prevent errors due to geoscheme updates.

\paragraph{Evaluation Metrics} We process generated text to map it to one of the geographical sub-regions or a policy violation case and then compare it with true labels by mapping country information to geographical regions, which gives us classification accuracy as a quantitative metric for measuring success. Since this is a typical classification problem, we also inspect the confusion matrix to locate sub-regions with more errors.

\paragraph{Economic disparities} For \dollarstreet{}, we also have data available for the monthly income of the family corresponding to each image. We use this information to understand differences in performance across economic groups by looking at region-specific normalized income quartiles.

\begin{figure}[t]
    \centering
    \begin{minipage}{0.45\textwidth}
        \small
        \centering
        \begin{tabular}{lcc}
            \toprule
            & \textbf{\gpt4v{}} & \textbf{\llava{}} \\
            \midrule
            \dollarstreet{} & $36.28$ & $\mathbf{36.83}$ \\
            \dataset{} & $56.31$ & $\mathbf{78.05}$ \\
            \marvl{} & $\mathbf{41.59}$ & $19.14$ \\
            \bottomrule
        \end{tabular}
        \caption{\llava{} matches or outperforms \gpt4v{} on two of three datasets. Human accuracy on a \dataset{} subset is 47.63\%.}
        \label{tab:task1-overall}
    \end{minipage}
    \hfill
    \begin{minipage}{0.45\textwidth}
        \centering
        \includegraphics[width=\textwidth]{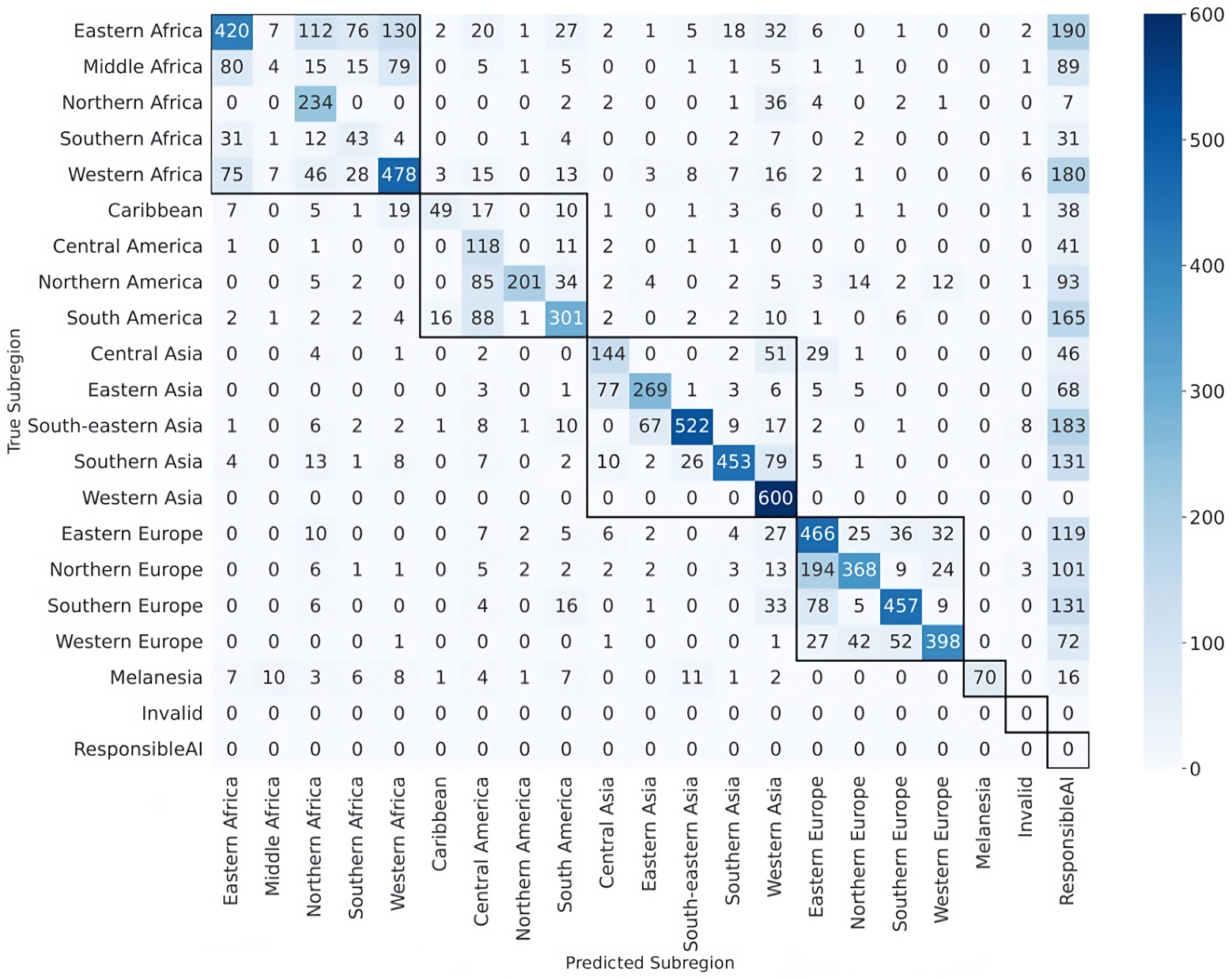}
        \caption{Confusion matrices for \gpt4v{} on the cultural awareness task for \dataset{} images. Accurate responses match the true subregion. Special labels include Invalid (no match or incomplete) and ResponsibleAI (policy violation). \textbf{Takeaway:} The model performs well, with a strong leading diagonal and $100\%$ accuracy for Western Asia (which covers Iran, Jordan, Lebanon, Oman, Palestine, Turkey).}
        \label{fig:conf-dalle-gpt}
    \end{minipage}
    \vspace{-1em}
\end{figure}

\paragraph{Human Baseline} $14$ annotators label images at the country, subregion, or continent level, with $1$ to $5$ guesses per image, accounting for varying familiarity with different regions. We first evaluate exact match accuracy at the country, then subregion, and finally continent levels. We also consider two cases: union (the correct answer appears at any level) and intersection (the correct answer appears at all levels). Table~\ref{tab:human-accuracy} shows that while country-level accuracy is low, it nearly doubles at each broader geographic level (Table~\ref{tab:user_scores}).

\subsection{Results}
We find similar trends across datasets for both models, with some variations across subregions.

\begin{figure}[t]
    \centering
    \includegraphics[width=0.48\textwidth]{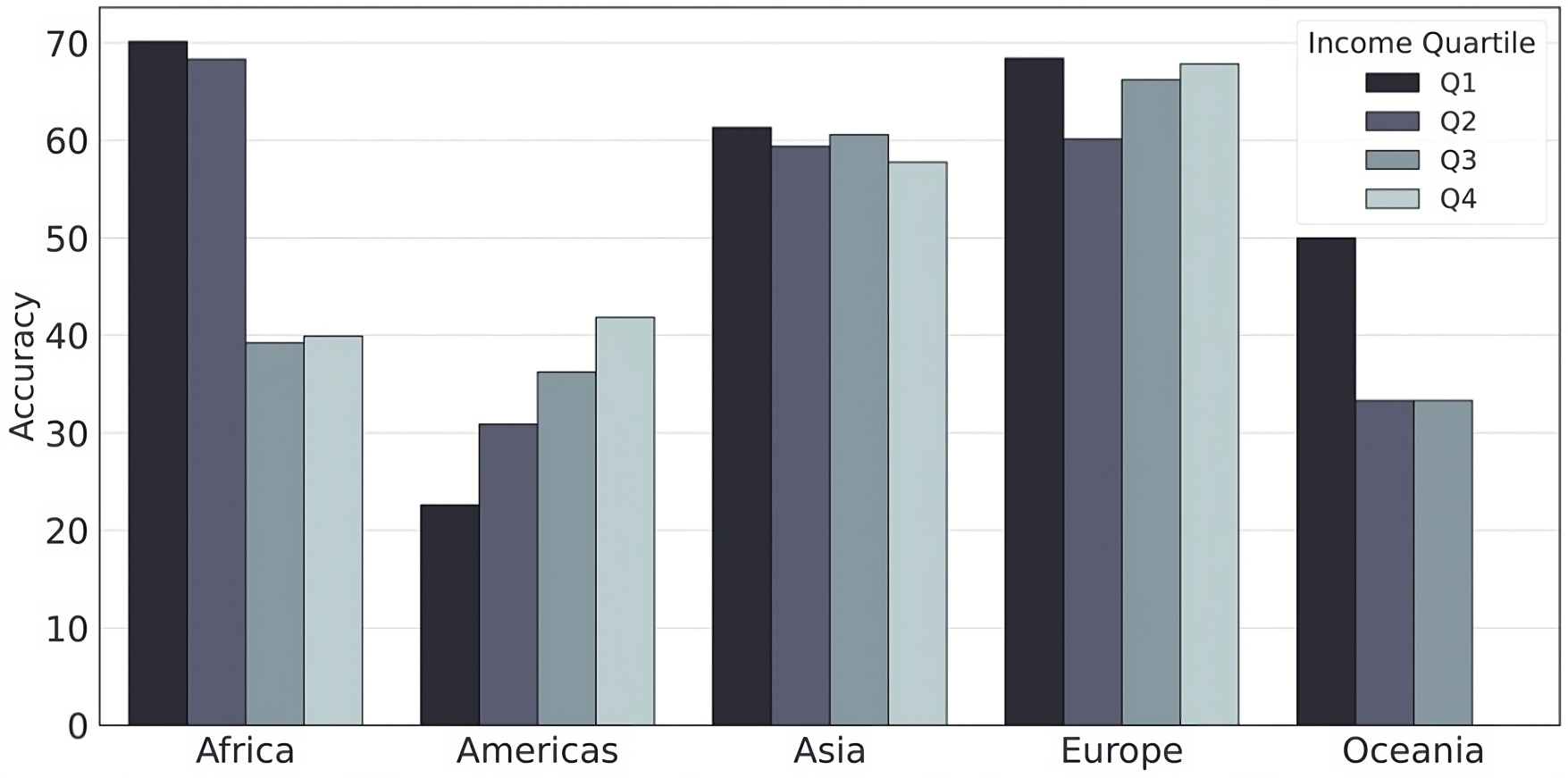}
    \caption{We normalize \dollarstreet{} income data into region-specific quartiles and plot accuracies for \gpt4v{}. \textbf{Takeaway:} Lower income quartiles (Q$1$, Q$2$) show higher accuracy in Africa and Asia, while higher quartiles (Q$3$, Q$4$) perform better in the Americas. In Europe, accuracy is similar across all quartiles.}
    \label{fig:gpt-income-disparity}
    \vspace{-1em}
\end{figure}

\paragraph{Overall comparison} \llava{} performs as good as \gpt4v{} on \dollarstreet{} and outperforms it significantly on the \dataset{} images (Table~\ref{tab:task1-overall}). This indicates that \llava{} \textit{may} have implicitly learned \textit{stereotypical} associations between regions and concepts because the \dataset{} images include such associations (Section~\ref{sec:artifact-extraction}). However, on the \marvl{} data, \llava{} performs about as good as random guessing. This \textit{may} be because \marvl{} covers specific indigenous concepts that the model may not have seen before.

\paragraph{Subregion Level Analysis} \gpt4v{} performs well on \dataset{}, with a strong leading diagonal indicating many correct predictions (Figure~\ref{fig:conf-dalle-gpt}). However, it often provides no answer due to content policy violations. Notably, both models accurately predict all Western Asian images (Figure~\ref{fig:conf-dalle-both}). \llava{} tends to default to South America for incorrect answers, while \gpt4v{} defaults to policy violations. Similar trends are observed in other datasets (Figures~\ref{fig:conf-dollar-both}, \ref{fig:conf-marvl-both}).

\paragraph{Economic Disparity} Using income data from \dollarstreet{}, we group results by normalized income quartiles across Africa, Asia, Americas, Europe, and Oceania (\gpt4v{} - Figure~\ref{fig:gpt-income-disparity}, \llava{} - Figure~\ref{fig:llava-income-disparity}). Performance is better for lower-income quartiles in Africa and Asia, while it improves with higher-income groups for the Americas. This \textit{might} indicate that the model defaults to associating Africa with poorer contexts and America with wealthier ones. For Europe, performance remains consistent across all quartiles.

\section{Extracting Implicit Associations of Cultures and Artifacts (Task 2)}
\label{sec:artifact-extraction}

We propose to extract cultural artifacts (material items) from the generated images to identify the implicit associations the models may use for Task~$1$. We find associations that are usually stereotypical (and not truly representative) for the relevant countries. This provides a better understanding of the models, enabling us to develop our approach for Task $3$ for cultural adaptation of images.

\begin{figure}[t]
    \centering
    \includegraphics[width=0.45\textwidth]{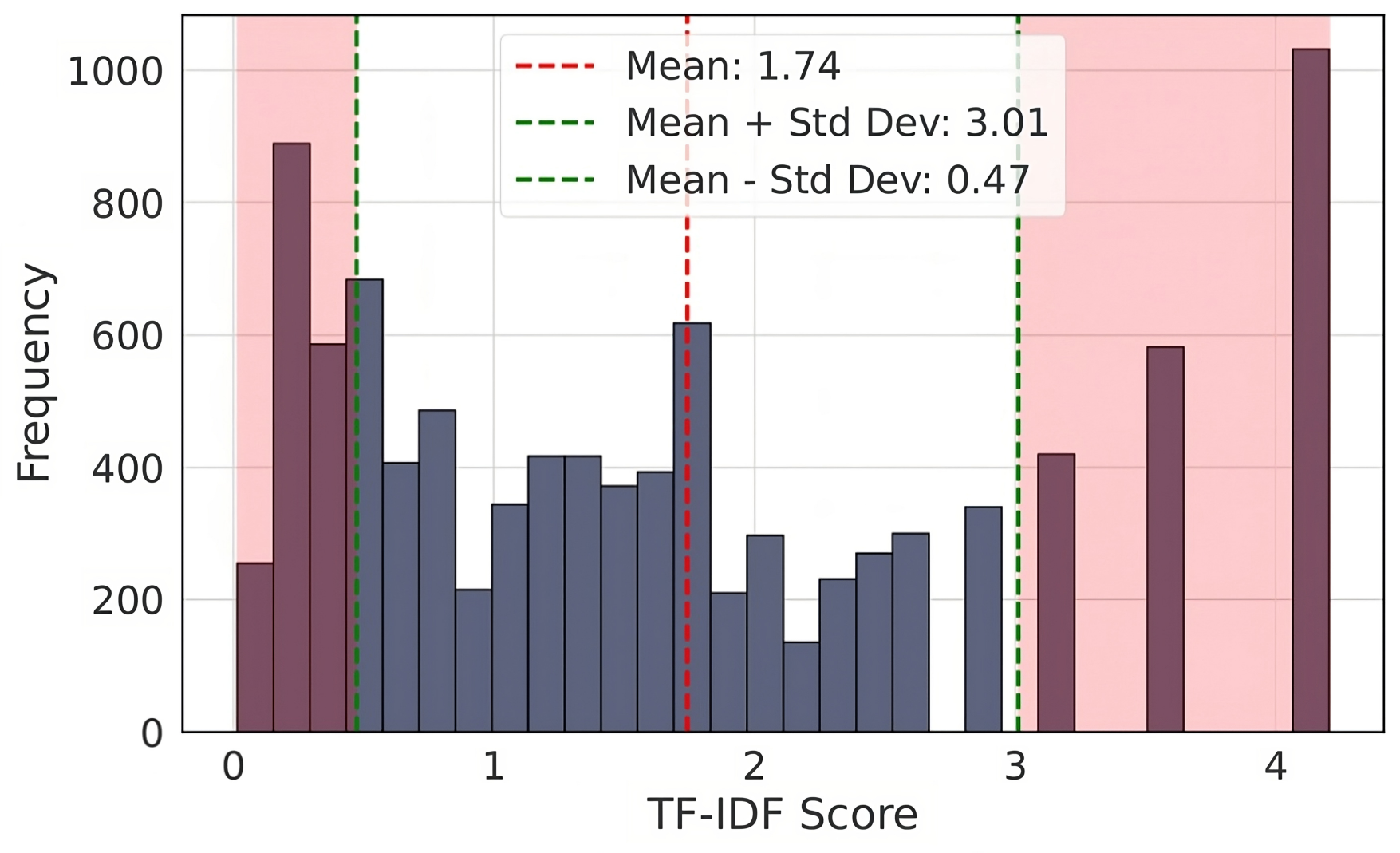}
    \caption{We score each artifact based on its likelihood of co-occurrence for a country. Scores outside the mean and standard deviation range (red) indicate frequent co-occurrences, representing implicit (potentially \textit{stereotypical}) associations.}
    \label{fig:tfidf-dist}
    \vspace{-1em}
\end{figure}

\subsection{Methods}
We use \gpt4v{} \texttt{vision-preview} for open vocabulary object detection \cite{alireza-2020-open-vocab}, using a detailed prompt (Figure~\ref{fig:object-detection-prompt}) to extract information about concept classes in \dataset{} images, including descriptions, color,\footnote{This refers to object and person appearance, not race.} and person count. \gpt4v{}'s strong instruction-following capabilities result in nearly perfect JSON outputs, which we lightly post-process and summarize using GPT-$4$ \texttt{turbo} (prompt in Figure~\ref{fig:object-detection-processing-prompt}). This process yields many unique associations for each country. In our initial experiments, \gpt4v{} significantly outperformed \llava{}, hence we report \gpt4v{} results.

\paragraph{Salient associations} To identify \textit{salient} artifacts that appear more frequently in one country than others (potentially \textit{stereotypical} associations), we follow an approach similar to ~\citet{jha-2024-visage}: compute the term frequencies of each artifact for each country and also compute document frequency as the number of times an artifact occurs across all countries, to calculate a \texttt{tf-idf} score by multiplying term frequency and the inverse of the document frequency. We then perform a qualitative evaluation of outliers from the distribution of these scores.

\begin{figure*}[t]
    \centering
    \includegraphics[width=\textwidth]{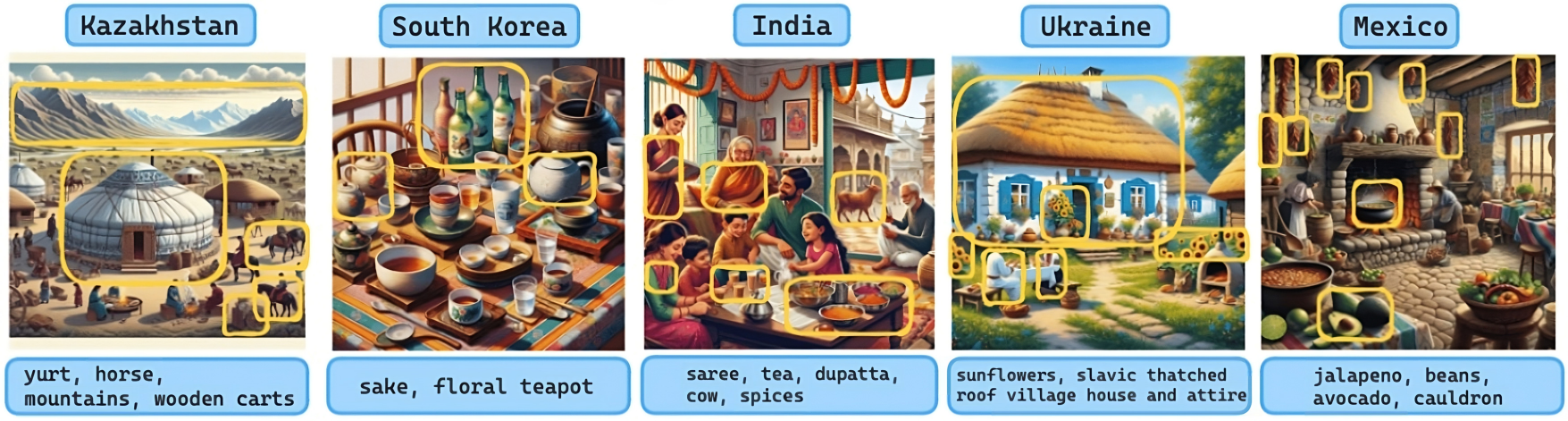}
    \caption{We identify more than $18{,}000$ unique cultural artifacts across all countries as part of our second task and then filter them to find salient ones (Table~\ref{tab:artifact-statistics}). This figure shows the strongest correlated artifacts for $5$ randomly picked countries. We include more such examples of country-object associations in the Appendix (Figure~\ref{fig:cultural-artifacts-all}).}
    \label{fig:culture-artifacts}
    \vspace{-1em}
\end{figure*}

\begin{figure}[t]
    \centering
    \includegraphics[width=.45\textwidth]{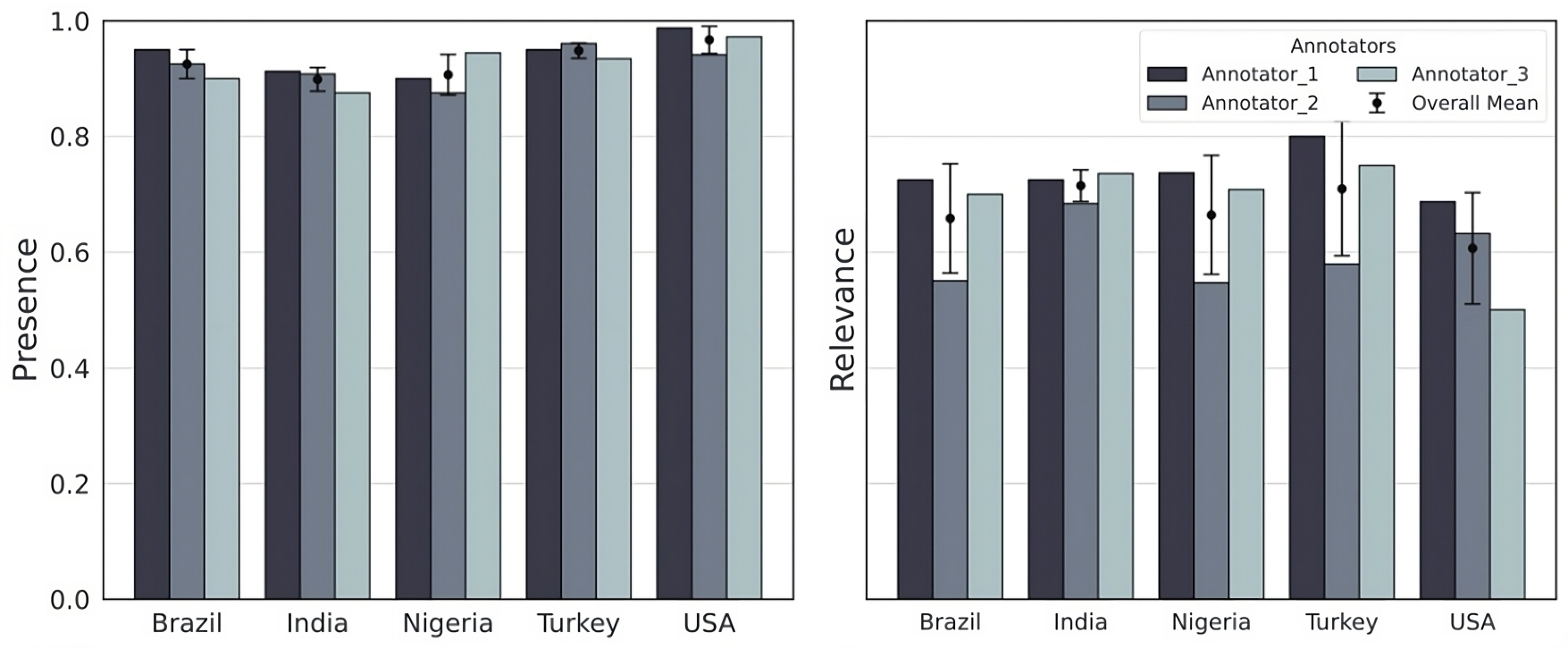}
    \caption{Human validation of a random sample of generated artifacts shows (1) low hallucination (high presence scores) and (2) more than a random fraction of artifacts usually display cultural relevance.}
    \label{fig:artifact-validation}
    \vspace{-1em}
\end{figure}

\paragraph{Evaluations} Extracting cultural artifacts is a novel task with no prior work or established metrics for quantitative evaluation. We explored several approaches, but each had limitations. As discussed before, our \dataset{} validation includes visual cues marked by annotators, consisting of names and bounding boxes. A simple metric could compare these names with objects extracted by \gpt4v{}, but annotators often provided descriptive labels (e.g., ``\textit{Mongol-looking structure}'') instead of specific terms (e.g., ``\textit{yurt}''), making semantic matching challenging. Bounding boxes were also imprecise, often covering multiple cues, limiting their usefulness. Given these challenges, we focus on salient association identification and qualitative evaluations. We sampled $100$ images and asked annotators to verify if ($1$) a cultural artifact is present and ($2$) if it is indeed culturally relevant. Over $90\%$ of the artifacts are deemed to be present, but only $60$--$70\%$ are marked relevant, indicating that not all salient associations are necessarily stereotypical (Figure~\ref{fig:artifact-validation}). We also compared human-labeled visual cues with model outputs, finding many similarities (Tables~\ref{tab:artifact-interesting},\ref{tab:artifact-human}). Future work could develop large-scale annotations for reference-based quantitative metrics.

\paragraph{Color Associations for Countries} We calculate the mean \texttt{RGB} vector for each \dataset{} image, then average them to get a global mean vector. We repeat this process at the country level and measure each country's distance from the global mean, identifying colors more strongly associated with specific countries across the three \texttt{RGB} dimensions.

\paragraph{Counting the Number of People} We observed that \dalle{} generates varying population densities across countries for identical prompts. To explore this, we use an object detection prompt (Figure~\ref{fig:object-detection-prompt}) to count people in each image, split into three buckets: less than $5$, $5$ to $10$, and more than $10$ people. We process related terms (e.g., people, person, man, woman) and aggregate country-level statistics to analyze population density distributions in the generated images. Annotators validate a random sample of these counts, finding general consistency, though they often do not reflect real-world population densities accurately.

\begin{figure}[t]
    \centering
    \vspace{-1em}
    \includegraphics[width=0.45\textwidth]{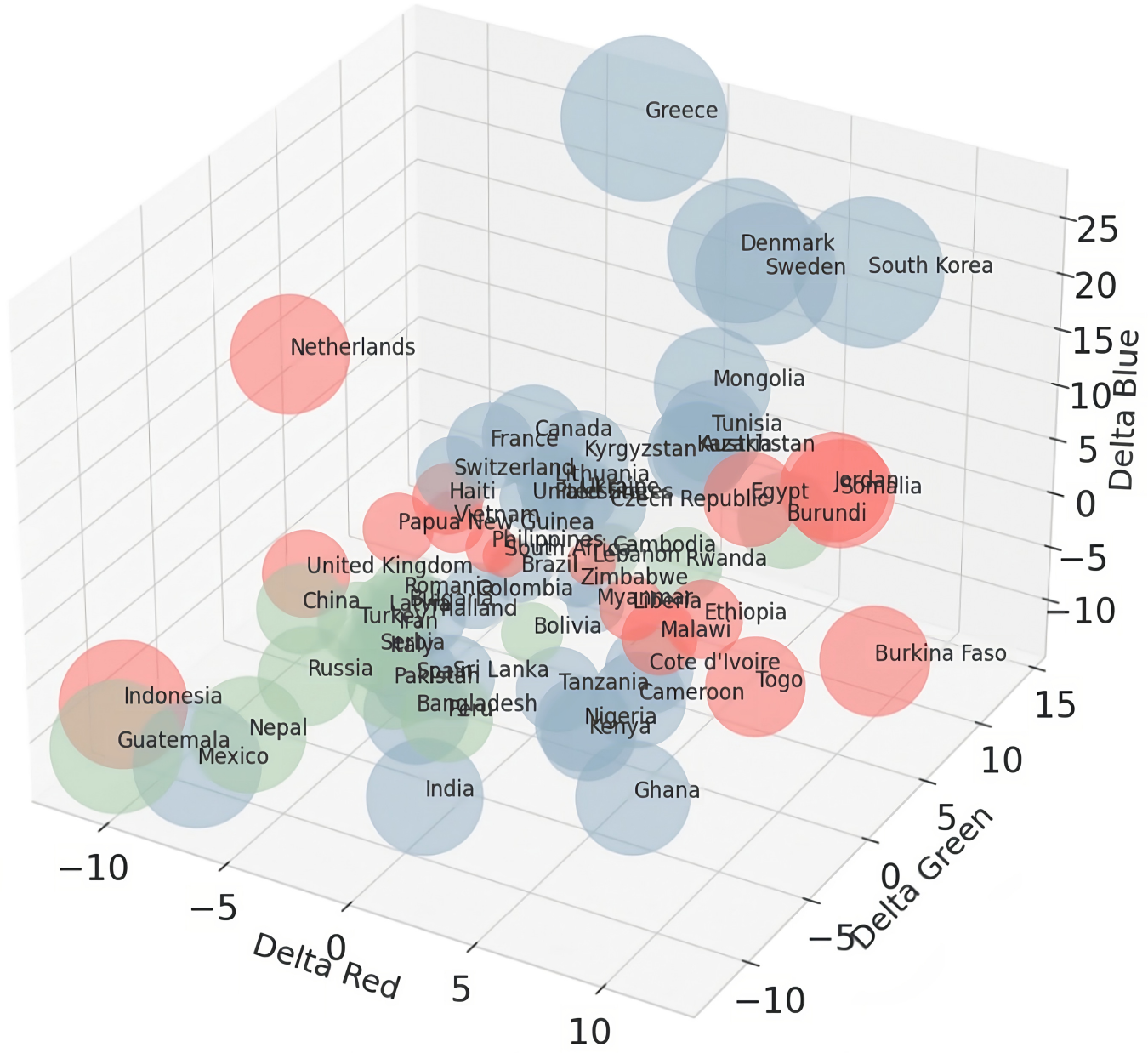}
    \vspace{-1em}
    \caption{We explore how countries are distributed on a color spectrum by first calculating a global average \texttt{RGB} vector and then defining deltas along each axes aggregated at the country level. \textbf{Takeaway}: We find interesting associations - Greece is strongly correlated with blue, and the Netherlands with red/orange.}
    \label{fig:rgb-deltas-3d}
    \vspace{-1em}
\end{figure}

\subsection{Results}
We analyze all the \dalle{} generated images to discover implicit associations between countries and cultural artifacts. Our method effectively surfaces implicit associations but also highlights challenges with stereotype reinforcement and demographic inaccuracies.

\paragraph{Artifact Associations} Our analysis shows that some cultural artifacts are strongly tied to specific countries, with certain artifacts exceeding one standard deviation from the mean \texttt{tf-idf} score (Figure~\ref{fig:tfidf-dist}). While these associations can offer cultural insights, they often reflect negative stereotypes, such as the over-representation of palm trees for tropical regions, which overlooks broader diversity. This suggests that while our method captures implicit associations, it also underscores the need to refine models to avoid reinforcing such stereotypes.

\paragraph{Color Associations} Models not only associate cultural artifacts with countries but also colors. In Figure~\ref{fig:rgb-deltas-3d}, the \texttt{RGB} delta values for several countries in \dataset{} fall outside the standard deviation. For instance, Greece is strongly associated with blue and the Netherlands with red, \textit{likely} due to recurring elements like blue seas in Greek images and red tulip fields in Dutch ones.

\begin{figure}[t]
    \centering
    \includegraphics[width=0.45\textwidth]{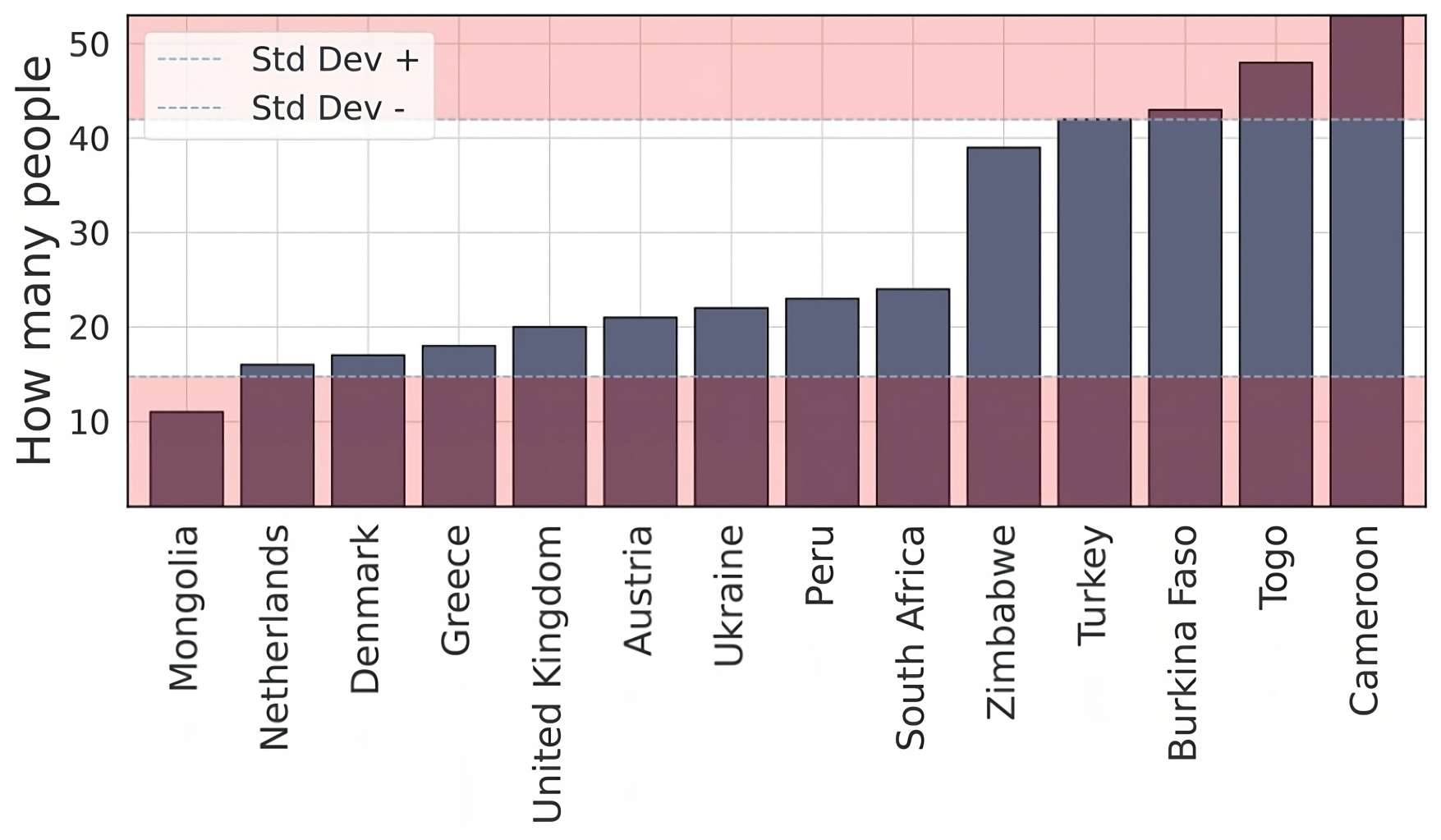}
    \caption{We explore people count associations being made by \dalle{} and \gpt4v{}, and show some selected countries where generated images in \dataset{} have more than $10$ detected people. \textbf{Takeaway:} African countries typically fall into high-person-count buckets in our experiments.}
    \label{fig:selected-people-counter}
    \vspace{-1em}
\end{figure}

\paragraph{People-Count Associations} Most images fall into the extreme buckets (less than $5$ or more than $10$ people), with few in the middle (Figures~\ref{fig:people-counts},\ref{fig:selected-people-counter}), often misrepresenting actual population densities. In general, African countries tend to fall into high person-count buckets, while European ones are on the low person-count end, \textit{possibly} reflecting the model's perception of collectivist versus individualistic societies.

\section{Cultural Adaptation (Task 3)} 
\label{sec:cultureadapt}

We propose a method to edit a given image for a target culture by modifying the detected salient implicit associations between countries and artifacts.

\subsection{Methods}
Recent works on cultural translation \cite{simran-2024-image-thousand-words, li-zhang-2023-cultural, fung-2024-nclb} define different approaches for adapting images or text from one culture to another. 

\begin{figure*}[t]
    \centering
    \includegraphics[width=0.9\textwidth]{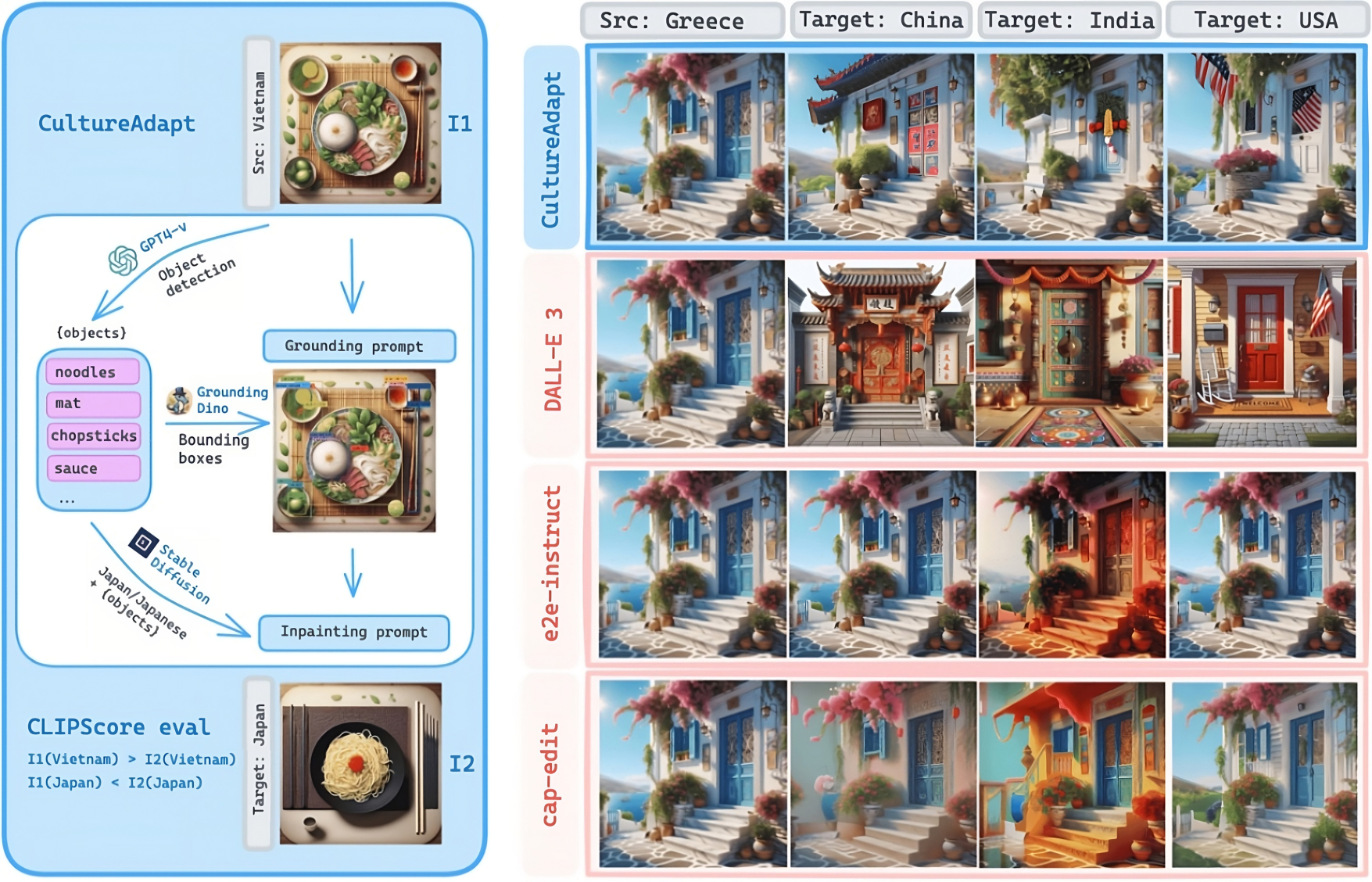}
    \caption{The \cultureadapt{} pipeline first identifies objects in an image using GPT-4V and grounds them with bounding boxes from \dino{}. These masks, along with a prompt containing object names and the target country, are used by Stable Diffusion for inpainting and cultural adaptation. \textbf{Takeaway:} Unlike \dalle{} (which generates a completely new image) or other editing methods \cite{simran-2024-image-thousand-words} that struggle with cultural adaptation, \cultureadapt{} makes precise, meaningful edits. Additional examples in the Appendix (Figure~\ref{fig:culture-adapt-edits}).}
    \label{fig:culture-adapt-vs-dalle}
    \vspace{-1em}
\end{figure*}

\paragraph{\cultureadapt{}} Our pipeline (Figure~\ref{fig:culture-adapt-vs-dalle}) uses \gpt4v{} for open-vocab object detection to extract implicit cultural associations from the source image. Next, we use \dino{}~\cite{shilong-2023-grounding-dino} to ground these objects with bounding boxes, which we convert into masks. We then create an inpainting prompt by adding the target country to the list of detected objects and use \sd{} $2$ \texttt{inpainting}~\cite{robin-2021-stable-diffusion} to edit\footnote{We acknowledge that the resulting image may represent common stereotypes because the underlying artifacts may be implicitly associated with a stereotypical view of the country.} the image by filling in the masked pixels. We evaluate our method using \clipscore{}~\cite{hessel-2021-clipscore} to measure image-country similarity (treating the name of the country as the caption) and cosine similarity of DINO-ViT~\cite{caron2021emerging}  embeddings for measuring structural preservation. Our choice of metrics is inspired from ~\citet{simran-2024-image-thousand-words}.

\paragraph{Modularity} Our pipeline is modular, allowing components to be easily swapped to improve performance over time. For example, we could use Tag2Text~\cite{huang-2023-tag2text} or RAM~\cite{zhang-2023-ram} for image captioning, extract object tags, and then use Grounded SAM~\cite{ren-2024-grounded-sam} for bounding boxes and segmentation masks. These can then be passed to inpainting models like Stable Diffusion 3~\cite{stable-diffusion-3} or MimicBrush~\cite{chen-2024-mimic-brush}.

\paragraph{Baseline} The closest related work is by~\citet{simran-2024-image-thousand-words}, which proposes three methods for image editing. We compare our approach with their two most relevant methods, providing qualitative examples (Figure~\ref{fig:culture-adapt-vs-dalle}) and quantitative evaluations (Table~\ref{tab:combined_similarity_comparison}). Additionally, we conduct a qualitative study assessing human preferences for layout preservation and cultural relevance changes (Figure~\ref{fig:edit-compare-validation}) for country pairs common to both studies. We also perform extensive statistical testing to compare structural similarity and editing success, using \clipscore{}-based metrics, for both approaches.

\paragraph{Evaluation} Let image \( I_1 \) correspond to country \( C_1 \) and \( I_2 \) be its adaptation for country \( C_2 \). The \clipscore{} for an image-country pair is denoted as \( S(I, C) \). We define two deltas: \vspace{-1em}

\begin{small}
\begin{align}
    \Delta_1 &= S(I_2, C_1) - S(I_1, C_1) \label{eq:delta1}\\
    \Delta_2 &= S(I_2, C_2) - S(I_1, C_2) \label{eq:delta2}
\end{align}
\end{small}

If $\Delta_1 < 0$ and $\Delta_2 > 0$, it indicates successful adaptation, where \( I_2 \) is closer to \( C_2 \) than \( C_1 \). Our primary metric, \( M_1 \), tracks how often this condition is met. A secondary metric, \( M_2 \), compares \( \Delta_2 - \Delta_1 \) to evaluate success, though it's less ideal for cases where \( \Delta_1 \) is positive. We also use another metric, \texttt{SSIM}, to measure structural similarity between \( I_1 \) and \( I_2 \) via DINO-ViT embeddings. A good edit performs well across editing and similarity metrics.

\subsection{Results}

Empirical findings indicate that our method works well both qualitatively and quantitatively.

\paragraph{Qualitative comparisons} In Figure~\ref{fig:culture-adapt-vs-dalle}, we show \cultureadapt{} applied to a randomly selected image, for adapting from Greece to China, India, and the USA. By visually contrasting with results from \dalle{}, we see that our approach preserves structural similarity better, and by comparing with the baselines \texttt{e2e-instruct} and \texttt{cap-edit}, we demonstrate our method's ability to make meaningful edits. Figure~\ref{fig:culture-adapt-edits} includes more examples.

\begin{figure}[t]
    \centering
    \includegraphics[width=0.45\textwidth]{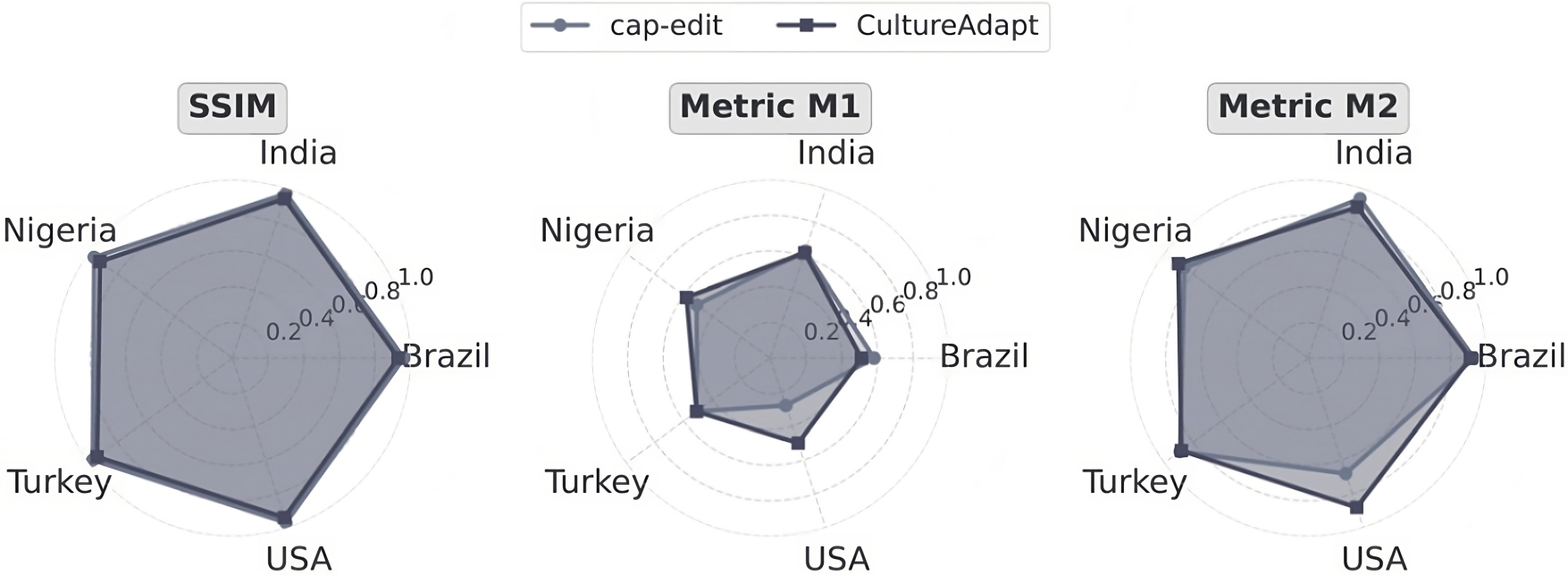}
    \caption{Our method performs better than other approaches in terms of editing metrics while still maintaining comparable structural similarities.}
    \label{fig:radar}
\end{figure}

\paragraph{Comparison with baseline} We compare \cultureadapt{} with \texttt{cap-edit} (Figure~\ref{fig:radar}) for $20$ country pairs (Table~\ref{tab:combined_similarity_comparison}). Both methods produce images similar to the source, but \cultureadapt{} edits are overall more culturally relevant.

To test this empirically, we compare similarity scores using the Wilcoxon signed-rank test (Shapiro-Wilk: \(p = 5.99 \times 10^{-39}\), indicating non-normality). \texttt{cap-edit} has a slightly higher mean similarity \(0.97\) than \cultureadapt{} \(0.94\), with a statistically significant difference (\(p = 6.02 \times 10^{-215}\)) for (\(\alpha = 0.05\)). The bootstrapped $95$\% confidence interval for the mean difference is [$0.0298$, $0.0333$], which does not include $0$, supporting this result.

\begin{figure}[t]
    \centering
    \includegraphics[width=.45\textwidth]{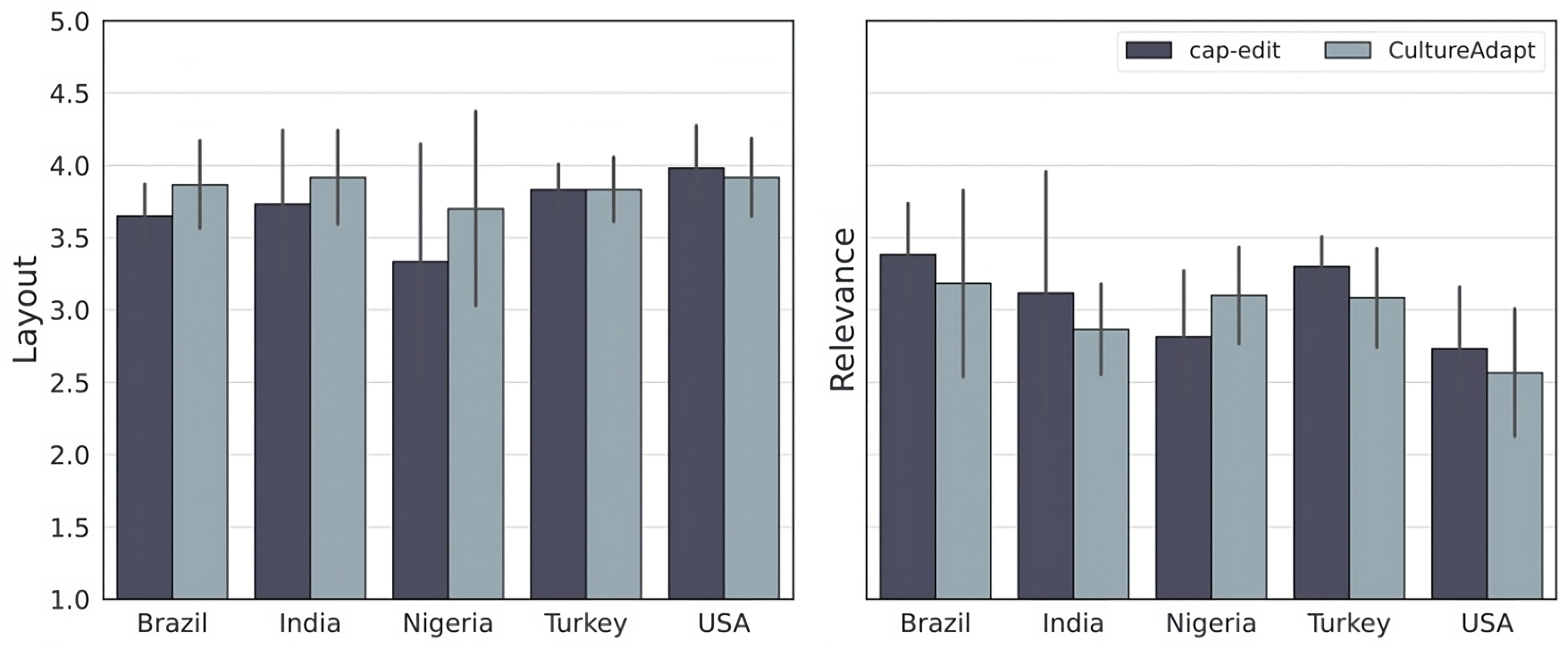}
    \caption{When asked to compare editing outputs from \cultureadapt{} to the existing baseline \texttt{cap-edit}, our annotators have nearly similar preferences for both methods in terms of (1) the ability of the methods to maintain layout similarity and (2) cultural relevance of the edited image to the target country.}
    \label{fig:edit-compare-validation}
\end{figure}

For editing metrics, \cultureadapt{} outperformed \texttt{cap-edit} in a statistically significant way. For $M_1$, $54\%$ of samples met the condition versus $50\%$ for \texttt{cap-edit} (\(p = 7.43 \times 10^{-5}\), McNemar's test). For $M_2$, \cultureadapt{} had a higher mean score ($3.11$ vs. $2.68$), with the difference significant per the Wilcoxon test (\(p = 4.39 \times 10^{-11}\)), as non-normality was confirmed by the Shapiro-Wilk test (\(p = 1.23 \times 10^{-16}\)). The bootstrapped $95\%$ confidence interval for $M_2$ differences supports our conclusion ([$-0.560$, $-0.307$]). In a human study of $100$ images, $3$ participants rated both methods equally preferable on structure preservation and cultural relevance (Figure~\ref{fig:edit-compare-validation}).

\begin{figure}[t]
    \centering
    \includegraphics[width=.45\textwidth]{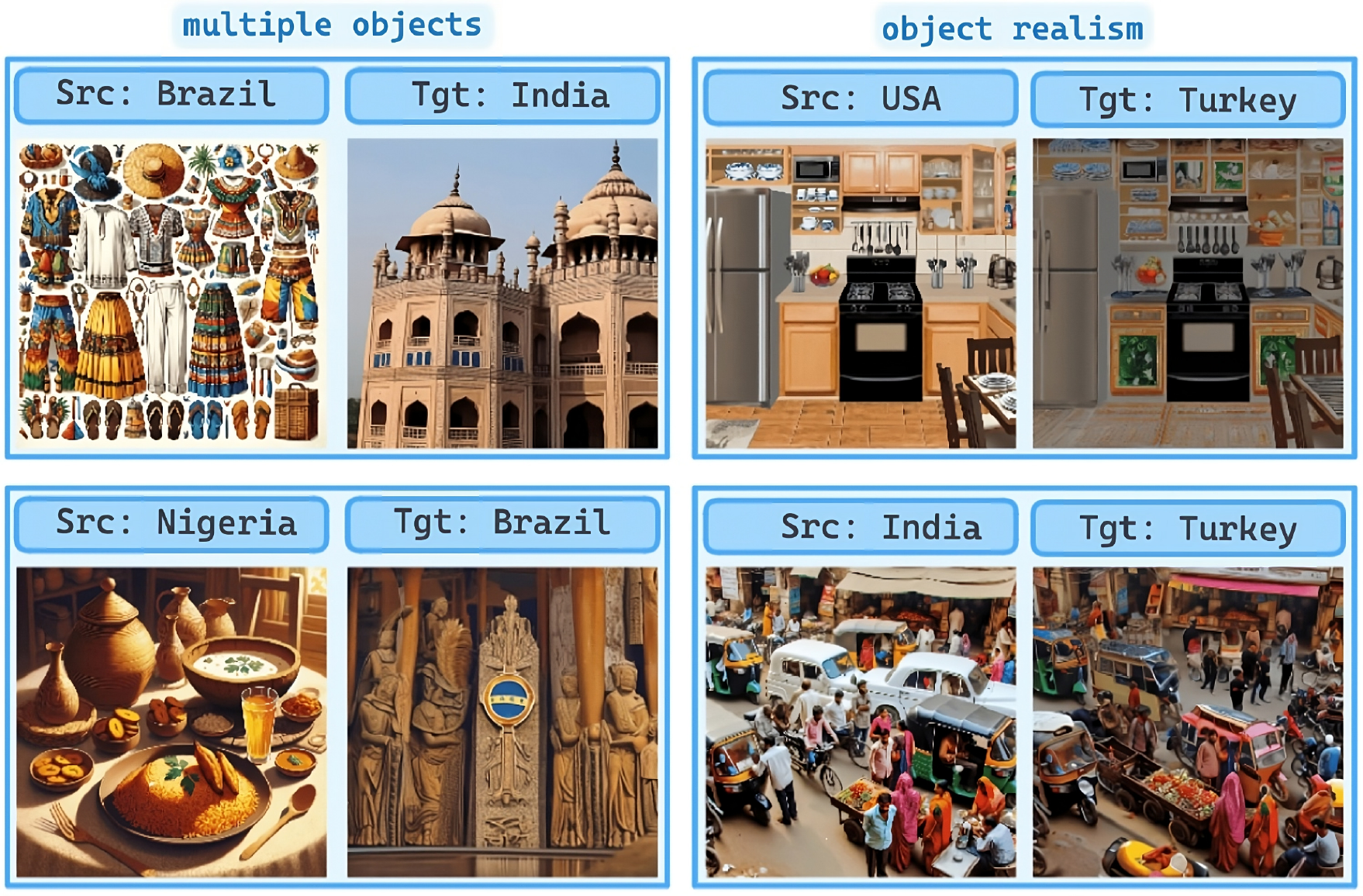}
    \caption{Common error cases with our method occur when there are multiple objects in the image or when the image is generated in a realistic style with \dalle{}.}
    \label{fig:error}
\end{figure}

\paragraph{Error Analysis} We identify two common error modes (Figure~\ref{fig:error}): ($1$) multiple similar objects to be edited in the source image lead to masks covering a significant portion of the image, often in an overlapping manner, resulting in major changes as the diffusion process needs to inpaint from scratch, and ($2$) when editing realistic \dalle{} images, the edited objects sometimes lose realism. Replacing \dino{} with segmentation models  (e.g., Grounded SAM) helps mitigate the first issue, but the second remains an open problem, as noted by others \cite{hall-2023-dig-in-a}. Other less obvious issues include not maintaining the correct count or orientation of objects and not generating human faces correctly, as the underlying diffusion model is trained with a privacy filter. 

\section{Related Work}
\label{sec:related-work}

The growing interest in culturally aware NLP has inspired various aspects of our research. 

\citet{li-zhang-2023-cultural} propose a data augmentation approach using semantic graphs to enhance cultural components in captions. However, their method often results in inconsistencies at object boundaries when cultural artifacts are copied and pasted into images. Similarly, \citet{simran-2024-image-thousand-words} formalize the task of image transcreation, but their pipelines can produce images that differ significantly from the source or not at all. Our \cultureadapt{} pipeline maintains image coherence by generating semantic masks with bounding boxes and using diffusion-based inpainting.

\citet{qiu-2024-valoreval} calculate co-occurrence statistics from image features, and \citet{jha-2024-visage} assign importance scores to attributes that frequently co-occur for identity groups. We build on these ideas to identify salient cultural artifacts likely to co-occur for a given country. \citet{liu-2024-culturally} provide a taxonomy for culturally aware NLP, from which we adopt terminology (cultural \textit{artifacts} and \textit{adaptation}). \citet{pouget-2024-no-filter} and \citet{hall-2023-dig-in-a} both explore the two real-world datasets, \marvl{} and \dollarstreet{}, specifically developing reliable metrics to measure geo-localization and object consistency across regions.

\section{Conclusion}
\label{sec:conclusion}
This study addresses the critical need for cultural awareness in Large Multimodal Models by introducing a comprehensive framework to evaluate and enhance their cultural competence. We create a large-scale, culturally diverse dataset of $9{,}935$ images across $67$ countries and $10$ concept classes, facilitating benchmarking of LMMs on cultural awareness tasks. Further, we introduce an artifact extraction task to identify over $18{,}000$ artifacts that co-occur frequently with these countries, revealing significant insights into the implicit cultural associations encoded in these models. We also propose \textsc{CultureAdapt}, a pipeline to adapt images across cultural contexts with fine-grained edits. Overall, this work emphasizes the importance of developing culturally sensitive AI systems and provides a foundational benchmark for future research toward improvement in cultural representation.

\section*{Limitations}
\label{sec:limitations}

Our experiments use a specific version of a closed-source API model, but later versions usually offer similar or improved performance at lower costs. The effectiveness of our cultural adaptation approach depends on its components, particularly object tag extraction, which currently works well when using \gpt4v{}. Open-source alternatives like RAM and Tag2Text would become more viable over time for this task. 

Our method could be extended for data augmentation to improve cultural awareness in models, though this falls outside the scope of our current work. Additionally, exploring multilingual prompts to analyze associations across languages would be interesting but is not our focus here. Since we use \sd{} for inpainting, we cannot fully prevent the stereotypical generations. These challenges point to the need for more culturally nuanced models, as discussed in recent works \cite{sung-2023-trak, hall-2023-dig-in-a, li-2024-culturepark}.

\section*{Acknowledgements}
\label{sec:ack}
This project was generously supported by the National Science Foundation under grant IIS-2327143 and by the Microsoft Accelerate Foundation Models Research (AFMR) grant
program.

\bibliography{anthology, custom}
\appendix
\clearpage
\onecolumn

\section{Appendix}
\label{sec:appendix}
\addcontentsline{toc}{section}{Appendix}

\begin{center}
    \textbf{Table of Contents}
\end{center}

\begin{enumerate}
    \item \textbf{Dataset details} \dotfill \pageref{sec:data-info}
    \begin{enumerate}
        \item \dollarstreet{} concept classes
        \item \dollarstreet{} countries
        \item Concept classes for \texttt{\textbackslash dataset\{\}}
        \item Countries in our \texttt{\textbackslash dataset\{\}}
        \item \marvl{} Country to Language Mappings
    \end{enumerate}
    \item \textbf{Prompt details} \dotfill \pageref{sec:prompt-details}
    \begin{enumerate}
        \item \dataset{} generation
        \item Cultural Awareness classifier
        \item Object Detection
    \end{enumerate}
    \item \textbf{LLM hyperparameters} \dotfill \pageref{sec:llm-hyperparameters}
    \begin{enumerate}
        \item Generation settings
        \item Computation budget
    \end{enumerate}
    \item \textbf{Human Study - the Annotators} \dotfill \pageref{sec:human-study-annotator-details}
    \begin{enumerate}
        \item Annotator Demographics
        \item How many studies do we have
        \item Total number of annotations
    \end{enumerate}
    \item \textbf{Human Study - the Interfaces} \dotfill \pageref{sec:human-study-interface-details}
    \begin{enumerate}
        \item General Instructions for Study 1 and 2
        \item Study 1: Verify the quality of generated images
            \begin{enumerate}
                \item Task 1 Instructions
            \end{enumerate}
        \item Study 2: Human performance on our cultural awareness benchmark
            \begin{enumerate}
                \item Task 2 Instructions
                \item Anonymized performance of individual Human Study participants
            \end{enumerate}
        \item Study 3 and 4: Artifact hallucination and cultural relevance
        \item Study 5: People count associations
        \item Study 6 and 7: Layout preservation and cultural relevance after editing
    \end{enumerate}
    \item \textbf{Additional Results} \dotfill \pageref{sec:additional-results}
    \begin{enumerate}
        \item Task 1 - Cultural awareness
            \begin{enumerate}
                \item Performance of \llava{} on income quartiles for \dollarstreet{}
                \item Confusion matrices for \llava{} and \gpt4v{} on \dataset{} images
                \item Confusion matrices for \llava{} and \gpt4v{} on \dollarstreet{} images
                \item Confusion matrices for \llava{} and \gpt4v{} on \marvl{} images
            \end{enumerate}
        \item Task 2 - Artifacts
            \begin{enumerate}
                \item Number of artifacts identified using \gpt4v{}
            \end{enumerate}
    \end{enumerate}
\end{enumerate}

\clearpage

\subsection{Dataset details}
\label{sec:data-info}

\paragraph{\dollarstreet{} concept classes (10)} car, family snapshots, front door, home, kitchen, plate of food, cups/mugs/glasses, social drink, wall decoration, and wardrobe.

\paragraph{\dollarstreet{} countries (63)} South Africa, Serbia, Indonesia, Brazil, Kenya, India, Nigeria, France, Kazakhstan, United States, Philippines, Mexico, Sri Lanka, Netherlands, Thailand, Colombia, Pakistan, China, Russia, Egypt, Iran, United Kingdom, Romania, Spain, Turkey, Ukraine, Italy, Czech Republic, Denmark, Ethiopia, Jordan, Burundi, Burkina Faso, Malawi, Somalia, Zimbabwe, Haiti, Cote d'Ivoire, Myanmar, Papua New Guinea, Liberia, Cambodia, Bangladesh, Rwanda, Nepal, Palestine, Tunisia, Cameroon, Bolivia, Ghana, Vietnam, Guatemala, Mongolia, South Korea, Kyrgyzstan, Lebanon, Tanzania, Switzerland, Sweden, Canada, Peru, Austria and Togo.

\paragraph{Concept classes (10) for \dataset{}} car, family snapshots, front door, home, kitchen, plate of food, cups/mugs/glasses, social drink, wall decoration, and wardrobe.

\paragraph{Countries in our \dataset{} (67)} Austria, Bangladesh, Bolivia, Brazil, Bulgaria, Burkina Faso, Burundi, Cambodia, Cameroon, Canada, China, Colombia, Cote d'Ivoire, Czech Republic, Denmark, Egypt, Ethiopia, France, Ghana, Greece, Guatemala, Haiti, India, Indonesia, Iran, Italy, Jordan, Kazakhstan, Kenya, Kyrgyzstan, Latvia, Lebanon, Liberia, Lithuania, Malawi, Mexico, Mongolia, Myanmar, Nepal, Netherlands, Nigeria, Pakistan, Palestine, Papua New Guinea, Peru, Philippines, Romania, Russia, Rwanda, Serbia, Somalia, South Africa, South Korea, Spain, Sri Lanka, Sweden, Switzerland, Tanzania, Thailand, Togo, Tunisia, Turkey, Ukraine, United Kingdom, United States, Vietnam, Zimbabwe

\paragraph{\marvl{} Country to Language Mappings} 
In the context of the \marvl{} dataset, various languages are mapped to specific sub-regions based on the countries where these languages are predominantly spoken. The mapping is as follows:

\label{marvl-labels}
\begin{itemize}
    \item \texttt{``id''}: The language code for Indonesian, which is primarily spoken in \textbf{Indonesia}, corresponds to the \textbf{South-eastern Asia} sub-region.
    \item \texttt{``sw''}: The language code for Swahili, used in countries such as \textbf{Tanzania}, \textbf{Kenya}, and \textbf{Rwanda}, is mapped to the \textbf{Eastern Africa} sub-region.
    \item \texttt{``ta''}: The language code for Tamil, spoken in \textbf{India} and \textbf{Sri Lanka}, is associated with the \textbf{Southern Asia} sub-region.
    \item \texttt{``tr''}: The language code for Turkish, which is the official language of \textbf{Turkey}, falls under the \textbf{Western Asia} sub-region.
    \item \texttt{``zh''}: The language code for Chinese, predominantly spoken in \textbf{China}, is linked to the \textbf{Eastern Asia} sub-region.
\end{itemize}

\begin{table}[htbp]
    \scriptsize
    \centering
    \caption{Dataset Statistics}
    \label{tab:statistics-sub-regions}
    \resizebox{\textwidth}{!}{
    \begin{tabular}{lcccccccccc}
        \toprule
        \textbf{Sub-region} & \textbf{Eastern Africa} & \textbf{Eastern Asia} & \textbf{South-eastern Asia} & \textbf{Southern Asia} & \textbf{Western Asia} & \textbf{Caribbean} & \textbf{Central America} & \textbf{Central Asia} & \textbf{Eastern Europe} & \textbf{Melanesia} \\
        \midrule
        \textbf{\marvl{}} & 875 & 1107 & 1091 & 924 & 917 & - & - & - & - & - \\
        \textbf{\dollarstreet{}} & 310 & 313 & 578 & 839 & 128 & 56 & 12 & 20 & 136 & 14 \\
        \textbf{\dalle{} Images} & 1052 & 438 & 840 & 742 & 600 & 160 & 176 & 280 & 741 & 147 \\
        \midrule
        \textbf{Total} & 2237 & 1858 & 2509 & 2505 & 1645 & 216 & 188 & 300 & 877 & 161 \\
        \bottomrule
        \\
        \toprule
        \textbf{Sub-region} & \textbf{Middle Africa} & \textbf{Northern Africa} & \textbf{Northern America} & \textbf{Northern Europe} & \textbf{South America} & \textbf{Southern Africa} & \textbf{Southern Europe} & \textbf{Western Africa} & \textbf{Western Europe} & \textbf{Total} \\
        \midrule
        \textbf{\marvl{}} & - & - & - & - & - & - & - & - & - & 4914 \\
        \textbf{\dollarstreet{}} & 107 & 81 & 317 & 51 & 447 & 60 & 223 & 262 & 183 & 4137 \\
        \textbf{\dalle{} Images} & 303 & 289 & 465 & 736 & 605 & 139 & 740 & 888 & 594 & 9935 \\
        \midrule
        \textbf{Total} & 410 & 370 & 782 & 787 & 1052 & 199 & 963 & 1150 & 777 & 18986 \\
        \bottomrule
    \end{tabular}}
\end{table}

\clearpage
\subsection{Prompt details}
\label{sec:prompt-details}

\subsubsection{\dataset{} generation}
\label{sec:appendixpromptexample}

\paragraph{Prompt for data generation}We use a simple template to prompt \dalle{} to generate images for a particular combination of country and category (Figure~\ref{fig:dalle-prompt}).

\begin{figure}[ht]
    \centering
    \begin{tcolorbox}[
        colback=mainbg,
        colframe=titlebg,
        title=\centering\footnotesize\dataset{} \texttt{Generation Prompt},
        fonttitle=\bfseries\color{fontcolor},
        coltitle=white,
        colbacktitle=titlebg,
        width=0.48\textwidth,
        boxrule=0.5mm,
        arc=3mm,
        top=1mm, bottom=1mm,
        left=1mm, right=1mm,
        boxsep=2mm
    ]
        \footnotesize
        \texttt{\color{fontcolor} A typical scene of \textbf{\{category\}} in \textbf{\{country\}}, culturally accurate and detailed.}
    \end{tcolorbox}
    \caption{We use a simple prompt that includes information about the concept class and the target country using a template, to generate our large scale dataset of \dalle{} images.}
    \label{fig:dalle-prompt}
\end{figure}

\subsubsection{Cultural Awareness classifier}

\paragraph{Prompt for classification of images} We use a simple prompt to generate names of subregions from models when provided an image.

\begin{figure}[h]
    \centering
    \begin{tcolorbox}[
        colback=mainbg,
        colframe=titlebg,
        title=\centering\footnotesize\texttt{Classification Prompt},
        fonttitle=\bfseries\color{fontcolor},
        coltitle=white,
        colbacktitle=titlebg,
        width=0.48\textwidth,
        boxrule=0.5mm,
        arc=3mm,
        top=1mm, bottom=1mm,
        left=1mm, right=1mm,
        boxsep=2mm
    ]
        \footnotesize
        \texttt{\color{fontcolor} Strictly follow the United Nations geoscheme for subregions. Which geographical subregion of the United Nations geoscheme is this image from? Make an educated guess. Answer in one to three words.}
    \end{tcolorbox}
    \caption{We use a simple prompt to classify images in the data, by generating subregion labels.}
    \label{fig:classifier-prompt}
\end{figure}

\subsubsection{Object Detection}
\paragraph{Prompt used for Object Detection with \gpt4v{}} We use a detailed prompt for \gpt4v{} to extract objects, colors and counts from images generated with \dalle{}.

\begin{figure}[h!]
    \centering
    \begin{tcolorbox}[
        colback=mainbg,
        colframe=titlebg,
        title=\centering\footnotesize\texttt{\gpt4v{} Object Detection Prompt},
        fonttitle=\bfseries\color{fontcolor},
        coltitle=white,
        colbacktitle=titlebg,
        width=\textwidth,
        boxrule=0.5mm,
        arc=3mm,
        top=1mm, bottom=1mm,
        left=1mm, right=1mm,
        boxsep=2mm
    ]
        \footnotesize
        \texttt{\color{fontcolor} Give me a json output of the items you see in this image in both the foreground and background. Output the objects as a JSON with two fields: 'relevant\_objects' for objects pertinent to the image category \textit{concept} and 'other\_objects' for all additional detected objects. Be as specific as possible. Within each field, for each detected object, include sub-fields describing object attributes like color, count, and anything else that is appropriate. For example, for buildings describe the architectural style in a sentence, for people describe clothing and headgear (if multiple colors and headgears are present, include the top three), for food items describe the exact type of food and include a brief recipe description, for pictures of rooms include objects in the background like mountains outside a window or paintings on the wall portraying something specific like a landmark or a particular type of scenery. For the counts of items, if the number of items is less than 10, give me exact numbers otherwise say more than 10.}
    \end{tcolorbox}

    \caption{We use a detailed prompt for \gpt4v{} to extract objects, colors and counts from images generated with \dalle{}.}
    \vspace{-1em}
    \label{fig:object-detection-prompt}
\end{figure}

\paragraph{Prompt used for processing generated objects} We use a detailed prompt for GPT-4 to process the dictionaries generated with the previous prompt into a simplified list along with some parsing rules to ensure correctness of the data structure.

\begin{figure}[h!]
    \centering
    \begin{tcolorbox}[
        colback=mainbg,
        colframe=titlebg,
        title=\centering\footnotesize\texttt{GPT-4 Processing Object Detection Outputs},
        fonttitle=\bfseries\color{fontcolor},
        coltitle=white,
        colbacktitle=titlebg,
        width=\textwidth,
        boxrule=0.5mm,
        arc=3mm,
        top=1mm, bottom=1mm,
        left=1mm, right=1mm,
        boxsep=2mm
    ]
        \footnotesize
        \texttt{\color{fontcolor} You will be provided a dictionary of items for the country \textbf{\{country\}} and concept \textbf{\{concept\}}. Summarize the dictionary into a list of comma-separated list of items with their respective colors. For example, \textit{[red apple, blue car, green tree, house with a red roof and tinted glasses]}. Strictly follow the output format requested. The dictionary is as follows: }
    \end{tcolorbox}
    \caption{We use a detailed prompt for GPT-4 to process the dictionaries generated with the previous prompt into a simplified list along with some parsing rules to ensure correctness of the data structure.}
    \vspace{-1em}
    \label{fig:object-detection-processing-prompt}
\end{figure}

\subsection{LLM hyperparameters}
\label{sec:llm-hyperparameters}

We discuss the generation settings we used for our experiments, and also the associated costs and hardware.

\subsubsection{Generation settings}

\begin{itemize}[nolistsep,noitemsep,leftmargin=*]
    \item \dalle{} images are generated for \texttt{vivid} and \texttt{natural} settings for \texttt{standard} quality and size $1024\times1024$
    \item GPT-4 and \gpt4v{} generations are obtained for temperature $= 0.7$, top\_p $= 0.95$, no frequency or presence penalty, no stopping condition other than the maximum number of tokens to generate, max\_tokens $= 300$.
    \item \llava{} generations are obtained for temperature $= 1.0$ and top\_p $= 1.0$, no penalties, and max\_tokens $= 128$. The reason for using a slightly higher temperature and top\_p is to have more consistent outputs. In our initial experiments, \llava{} did not perform as well in terms of following instructions at the same temperatue setting as \gpt4v{}.
    \item For \dino{}, we use \texttt{ShilongLiu/GroundingDINO} from Hugging Face and set both box and text thresholds to 0.25 for the grounded box generations.
    \item For \sd{}, we use \texttt{stabilityai/stable-diffusion-2-inpainting} from Hugging Face, and replace the autoencoder with \texttt{stabilityai/sd-vae-ft-mse}. We also use a \texttt{DPMSolverMultistepScheduler} for speeding up the generation process. We add \texttt{``intricate details. 4k. high resolution. high quality.''} to the end of our prompt to get high quality images.
\end{itemize}

\subsubsection{Computation budget}
\begin{itemize}[nolistsep,noitemsep,leftmargin=*]
    \item We spent about $\$800$ in total for \dalle{} generations. This was funded by a grant from Microsoft Azure.
    \item We spent about $\$700$ in total for \gpt4v{} \texttt{vision-preview} and GPT-4 \texttt{turbo} inference and across all experiments.
    \item For experiments with \llava{}, \sd{} and \dino{}, we used a single instance of a Multi-Instance A100 GPU with 40GB of GPU memory, 3/7 fraction of Streaming Multiprocessors, 2 NVIDIA Decoder hardware units, 4/8 L2 cache size, and 1 node. 
    \item Total emissions for API based models are estimated to be 25 kgCO$_2$ eq, of which 100 percents were directly offset by the cloud provider. Total emissions for our on-premise GPU usage is estimated to be less than 10 kgCO$_2$ eq. Estimations are conducted using the MachineLearning Impact calculator~\cite{lacoste-2019-quantifying}.
\end{itemize}

\clearpage

\subsection{Human Study - the Annotators}
\label{sec:human-study-annotator-details}

\paragraph{Annotator Demographics} All annotators have different demographic backgrounds (but are physically located in the USA currently) and are between 25-40 years old. Together, they are native to or have resided in more than countries and all 4 major regions from our dataset, and thus represent a strong sample of opinions. Roughly 40\% identify as female, and the rest as male. In terms of an education background, 40\% of the annotators are graduate students, whereas the rest includes working professionals from different backgrounds and also computer science faculty. In total, we have 14 annotators, recruited from different computer science labs at an university and also from a diverse set of social connections for this study. All the annotators have agreed to consent for using this data for research purposes. Our study qualifies for exemption from IRB as no PII is involved.

\paragraph{How many studies do we have}
\begin{enumerate}
    \item Human study to verify the quality of generated images 
    \item Human study to understand performance on our benchmark
    \item Human validation for checking if artifacts are hallucinated
    \item Human study to check if artifacts are culturally relevant
    \item Human study for verifying people count associations
    \item Human study for layout preservation after editing using CultureAdapt
    \item Human study for cultural relevance after editing using CultureAdapt
\end{enumerate}

\paragraph{Total number of annotations} To validate our dataset and human baseline on cultural awareness, we have $14$ people, each annotating $300$ samples, resulting in $4200$ annotations for one study and $8400$ in total. For each of the other three studies (excluding the studies for edits), we have three people annotating $100$ samples each for a study, resulting in $900$ annotations in total. Finally, for the two editing-related studies, we have three people each looking at three images per sample (source image, edited image using our method, edited image using another method) for $100$ samples, resulting in $900$ annotations per study or $1800$ in total. Combining all this, we have $8200+900+1800=10900$ annotation items to back the comprehensive quantitative evaluations we perform for each study in our paper.

\subsection{Human Study - the Interfaces}
\label{sec:human-study-interface-details}

\paragraph{General Instructions for Study 1, 2} We use \url{https://labelstud.io}~\cite{label-studio-2020} to perform our human study. For each annotator, we create 2 tabs corresponding to the 2 studies, and ask them to solve them in numerical order to avoid getting influenced from seeing true labels first from the second study. Time taken to complete the first study is usually 2 hours, and the time taken to complete the second study is typically 30 minutes. All annotators will be compensated for their time with a \$20 gift card upon completion of the task.

\subsubsection{Study 1: verify the quality of generated images }

\paragraph{Task 1 Instructions} For every image, you have to make atleast 1 guess for the geographical region label, along with atleast 1 corresponding clue. If you are not sure what the clue is, add a question mark symbol at the end of it - example, headgear? bread?). Do not reverse image search or look anything up. Answer only using your knowledge or instinct. You can try guessing sub-region/region if you are not sure about country. You can use the knowledge of the fact that the image is generated by an AI conditioned on the provided prompt above the image. Note that you do not have to be correct! \\

\begin{figure}[t]
    \centering
    \includegraphics[width=\textwidth]{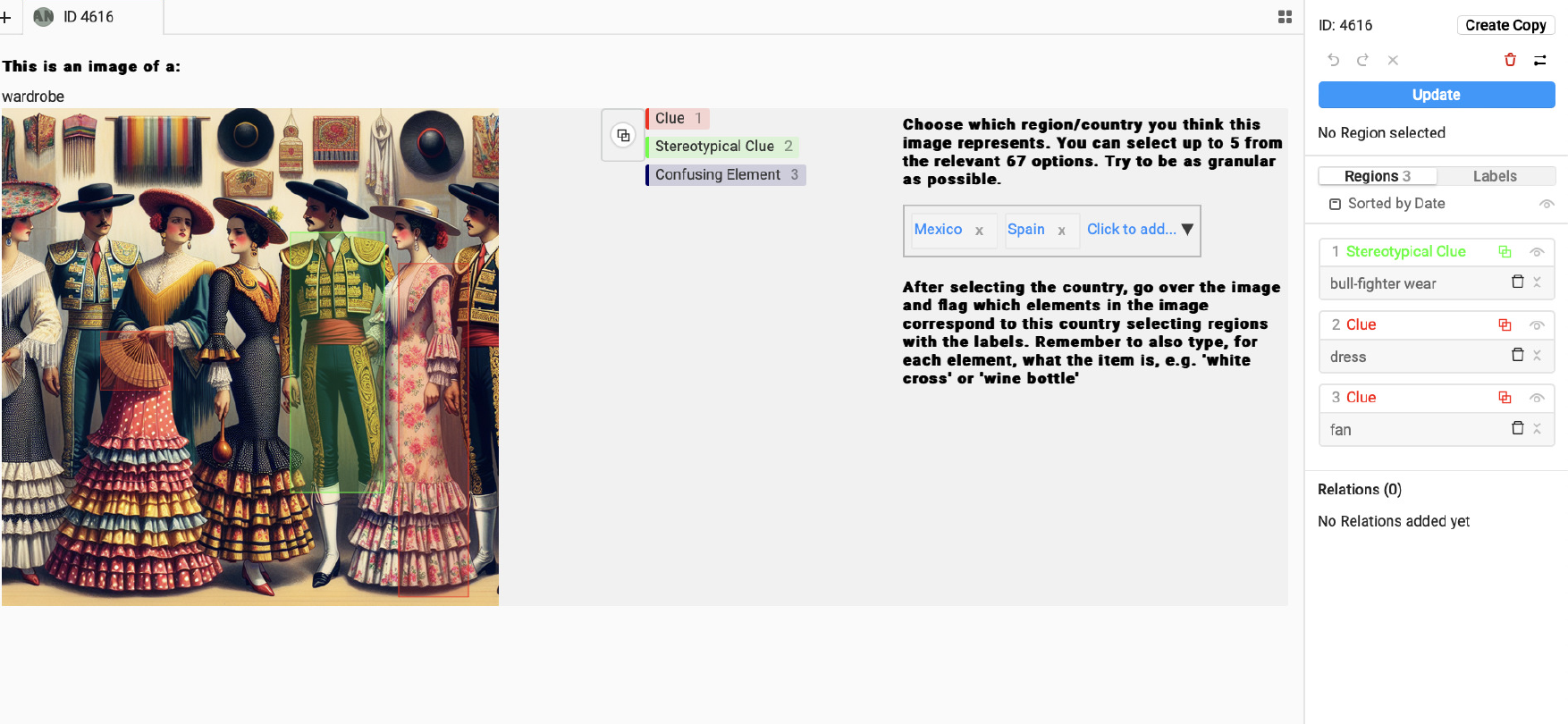}
    \caption{Annotation Interface for Study 1}
    \label{fig:annot-study-1}
    \vspace{-1em}
\end{figure}

\noindent Once the label is done, you need to add at least one bounding box somewhere in the image (it can be very specific and small or very broad or even the entire image) and then label that bounding box as either a clue or a stereotypical clue or a confusing element and then add a text description for the bounding box from the interface on the right (for example, a bounding box for a basket of baguettes can be a clue for France and the text description may be either something specific like ``baguette'' or something generic like ``bread?''). The difference between stereotypical clue and regular clue is that stereotypical would be something like baguettes for France or ``naan'' for India or specific clothing styles for some country whereas a regular clue is something that you are using to make your guess but you don’t know enough about your guess to know what stereotypical clues might be associated with it, for example, sand for island countries.

\subsubsection{Study 2: human performance on our cultural awareness benchmark }

\begin{figure}[h]
    \centering
    \includegraphics[width=\textwidth]{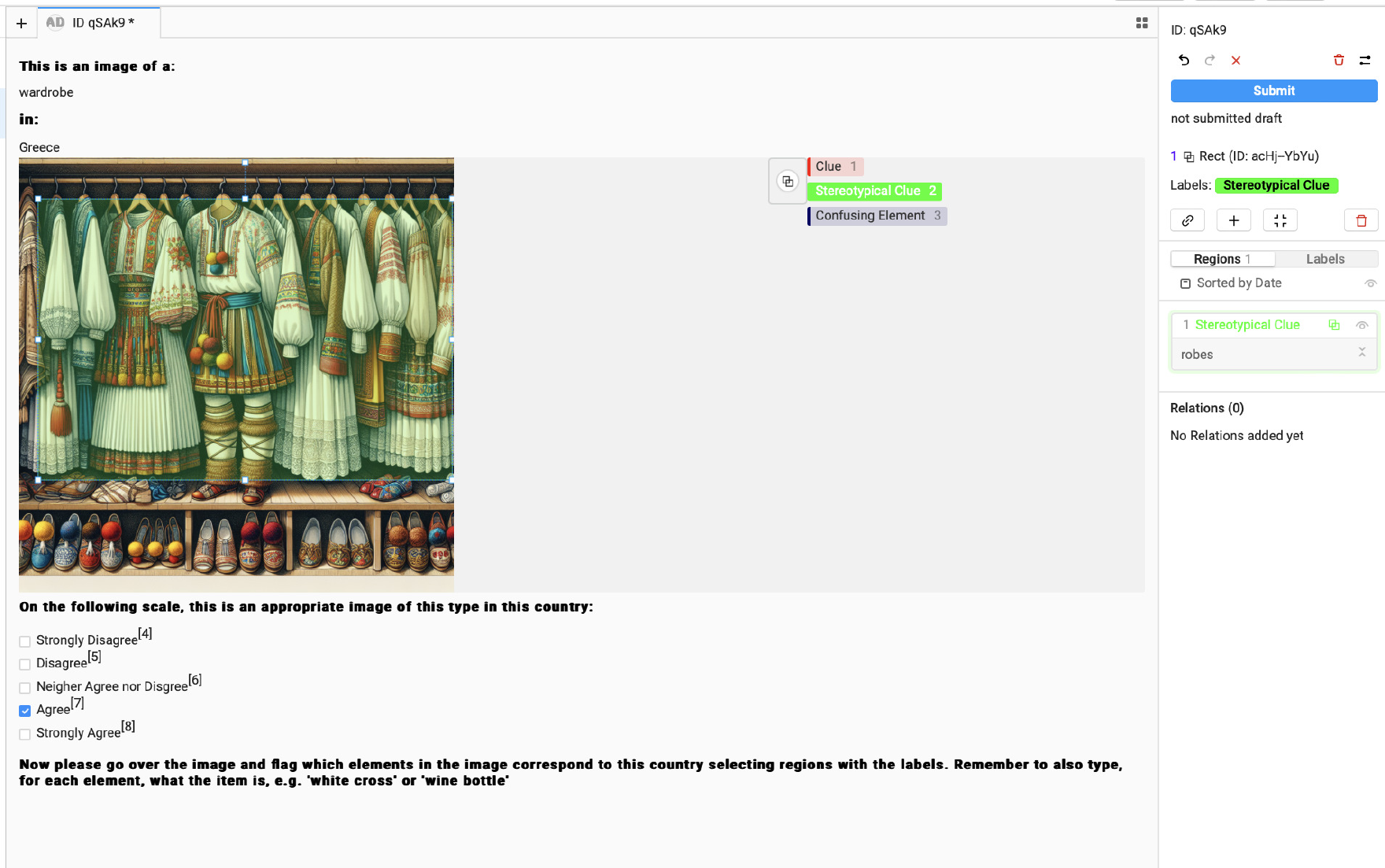}
    \caption{Annotation Interface for Study 2}
    \label{fig:annot-study-2}
    \vspace{-1em}
\end{figure}

\paragraph{Task 2 Instructions} For every image, select a single rating for ``appropriateness'' $=$ ``this image is one of the possible (stereo)typical representation of the mentioned category for the mentioned country''. Then select at least one clue corresponding to your rating, similar to the first study.

\paragraph{Performance of Human Study participants on the Cultural Awareness Task} We include anonymized performance statistics of each of our annotators to show differences in performance at the individual data point level for countries, subregions and continents.

\begin{table}[h!]
    \scriptsize
    \centering
    \begin{tabular}{cccc}
        \toprule
        \textbf{User} & \textbf{Country Level} & \textbf{Subregion Level} & \textbf{Continent Level} \\
        \midrule
        User 1 & 46.53 & 70.14 & 90.97 \\
        User 2 & 16.67 & 31.94 & 78.47 \\
        User 3 & 11.19 & 31.47 & 70.63 \\
        User 4 & 32.81 & 50.78 & 72.66 \\
        User 5 & 21.13 & 51.41 & 75.35 \\
        User 6 & 10.87 & 32.61 & 71.01 \\
        User 7 & 28.67 & 64.34 & 84.62 \\
        User 8 & 22.14 & 43.57 & 69.29 \\
        User 9 & 7.09 & 34.04 & 70.92 \\
        User 10 & 26.43 & 62.86 & 83.57 \\
        User 11 & 20.71 & 50.71 & 87.14 \\
        User 12 & 4.86  & 18.05 & 75.69 \\
        User 13 & 3.57  & 17.85 & 72.14 \\
        User 14 & 6.47  & 37.41 & 75.53 \\
        \bottomrule
    \end{tabular}
    \caption{User accuracies across country, subregion and continent levels, rounded to two decimal places. At the country level, accuracy varies from 46\% to about 4\%, so the subregion level accuracies are a more reliable indicator of performance even for humans. Continent is the most generic label, and has very high accuracies from all participants.}
    \label{tab:user_scores}
\end{table}

\subsubsection{Study 3, 4: artifact hallucination and cultural relevance }

\begin{figure}[h]
    \centering
    \includegraphics[width=\textwidth]{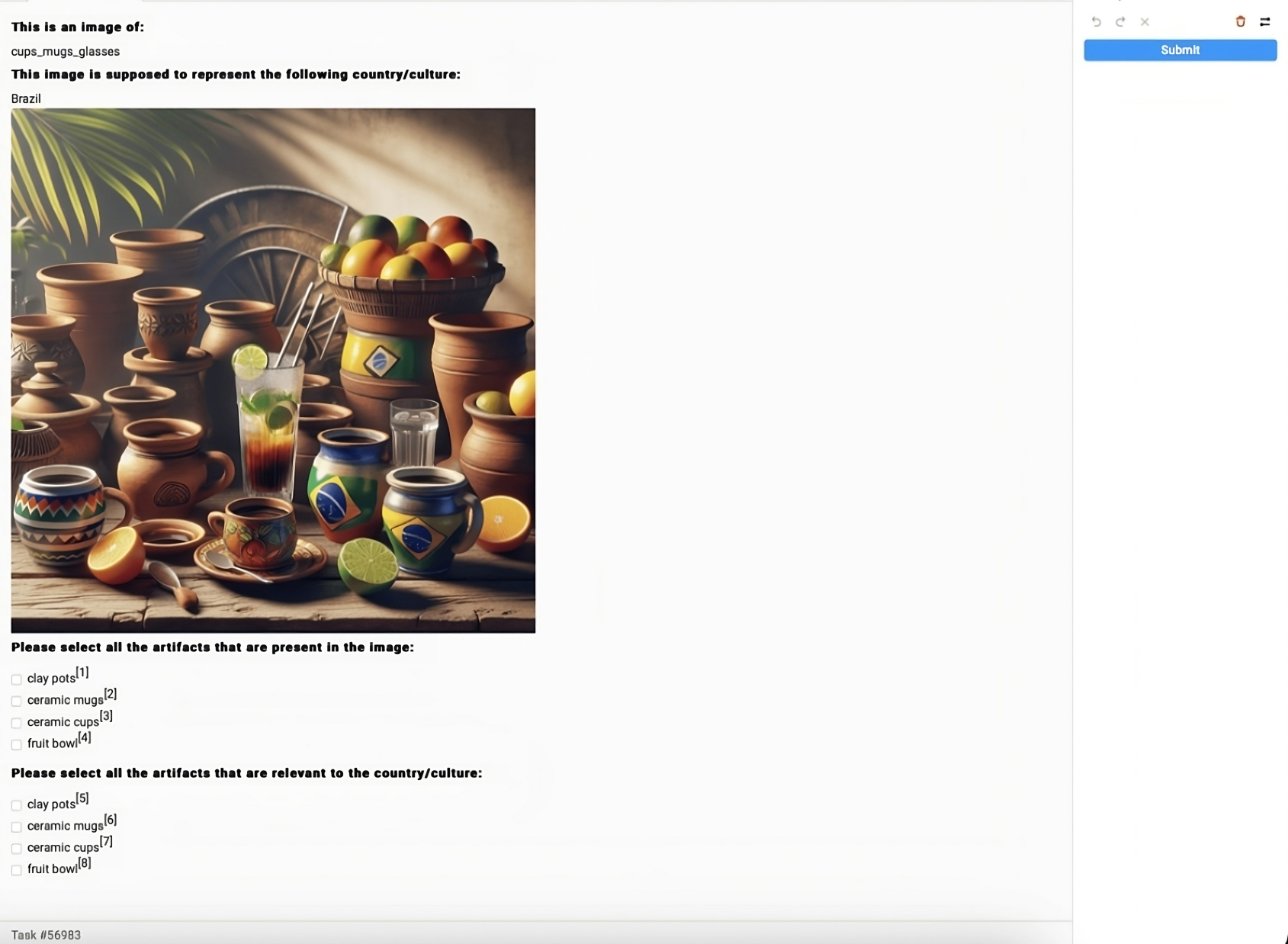}
    \caption{Annotation Interface for Study 3,4}
    \label{fig:annot-study-3-4}
    \vspace{-1em}
\end{figure}

We provide an interface (Figure~\ref{fig:annot-study-3-4}) that includes a multi-choice correct question answering setting, where we first ask a question about the artifacts that are present in the image and then ask if those artifacts are culturally relevant. For selecting these options, we sample randomly from all artifacts extracted for the accompanying image. For this study, we also provide the name of the concept class that the image is supposed to represent and the country, so that annotators have an idea about the object categories that they might be looking for. To account for cases where annotators might not be sure about any object and still choose one, in our analysis we only consider those cases where atleast two artifacts are marked as present. Responses to the second question is somewhat subjective as it depends on the cultural backgrounds of the annotators, but as we see in our analysis, the responses mostly agree and find more than half of the artifacts to be relevant.

\subsubsection{Study 5: people count associations }
We used an identical interface to the one for Study 3,4 to carry out this study, but instead of providing multiple correct options, we provide a direct True/False setting for annotators to answer if the count of people shown in the given image is approximately correct, i.e. falls in the correct count bucket (1-5, 5-10, more than 10). Our analysis reveals that people mostly agree with the count buckets generated, but on corroborating patterns with actual population statistics, the ordering of countries look slightly different, so we do not make any correlations about that in our analysis.

\subsubsection{Study 6, 7: layout preservation and cultural relevance after editing }

\begin{figure}[h]
    \centering
    \includegraphics[width=\textwidth]{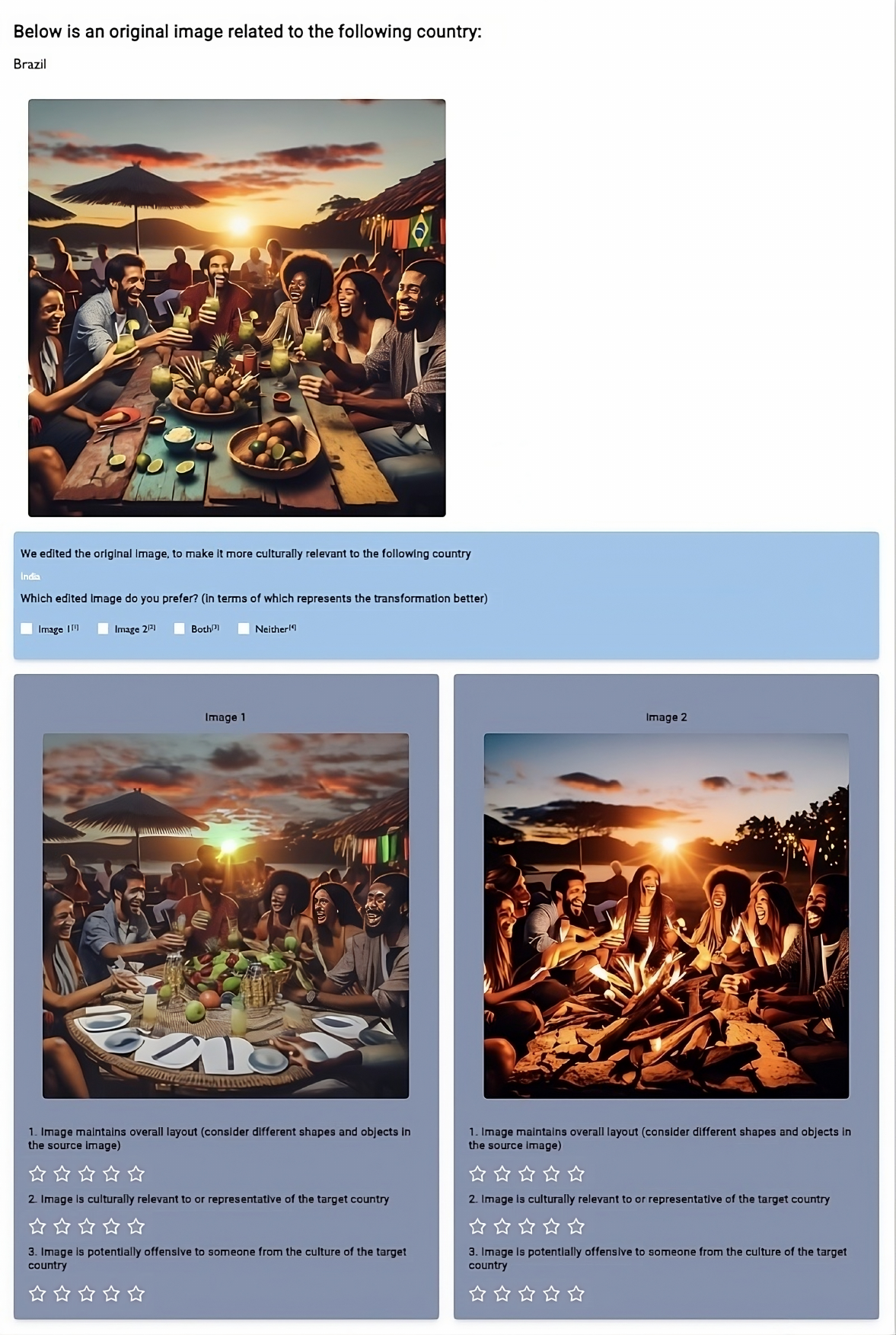}
    \caption{Annotation Interface for Study 6,7}
    \label{fig:annot-study-6-7}
    \vspace{-1em}
\end{figure}

We use an interface (Figure~\ref{fig:annot-study-6-7}) that shows the name of the source and target countries, the source image, and two edited images - one from our method \cultureadapt{} and one from \texttt{cap-edit}. We ask a somewhat subjective question regarding which image the annotator prefers, but responses vary widely for this question and do not correlate well across annotators, so we do not use these results for further analysis. For each of the edited images, we ask three questions - (1) if the edit maintains structural layout, (2) if the edit makes the image more culturally relevant to the target country and (3) if the edited image might be toxic for people from target country. All of these questions require a response on a Likert scale, which we design using a system of stars, where users can select anywhere between 1 and 5 stars. Less than 1\% of images are marked toxic across all annotations and annotators agree strongly on this metric. We report the analysis for the other two metrics in the main paper.

\subsection{Additional Results}
\label{sec:additional-results}

\subsubsection{Task 1 - Cultural awareness}

\paragraph{Income distribution performance for \llava{}} We show the performance of \llava{} on the income quartile distribution basis for \dollarstreet{}.

\begin{figure}[t]
    \centering
    \includegraphics[width=0.48\textwidth]{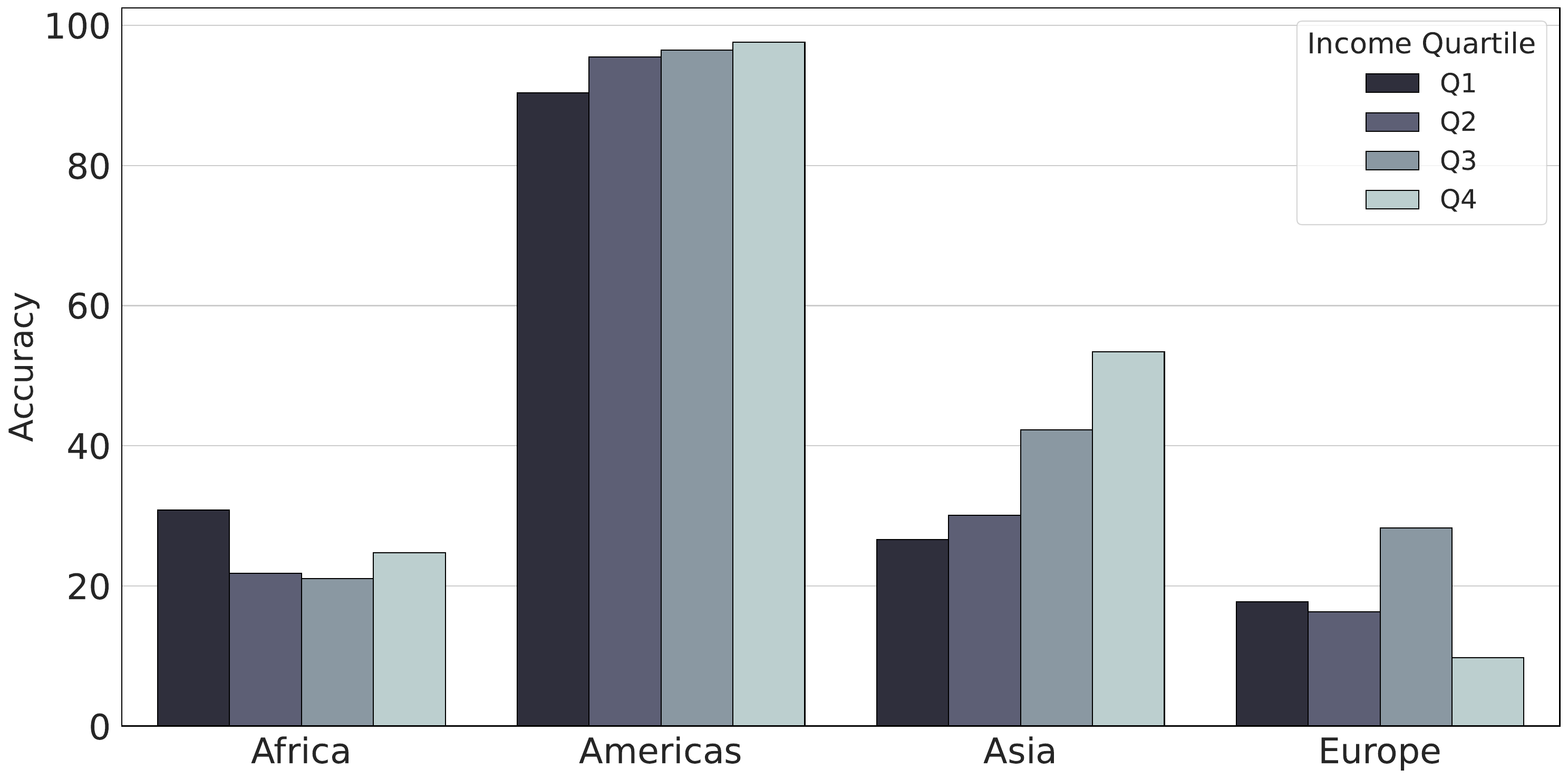}
    \caption{We normalize income data from \dollarstreet{} into region specific quartiles and plot  corresponding accuracies for \llava{}.}
    \label{fig:llava-income-disparity}
    \vspace{-1em}
\end{figure}

\paragraph{Confusion matrix set 1} We show confusion matrices for both \llava{} and \gpt4v{} on \dataset{} images.
\begin{figure*}[t]
    \centering
    \includegraphics[width=\textwidth]{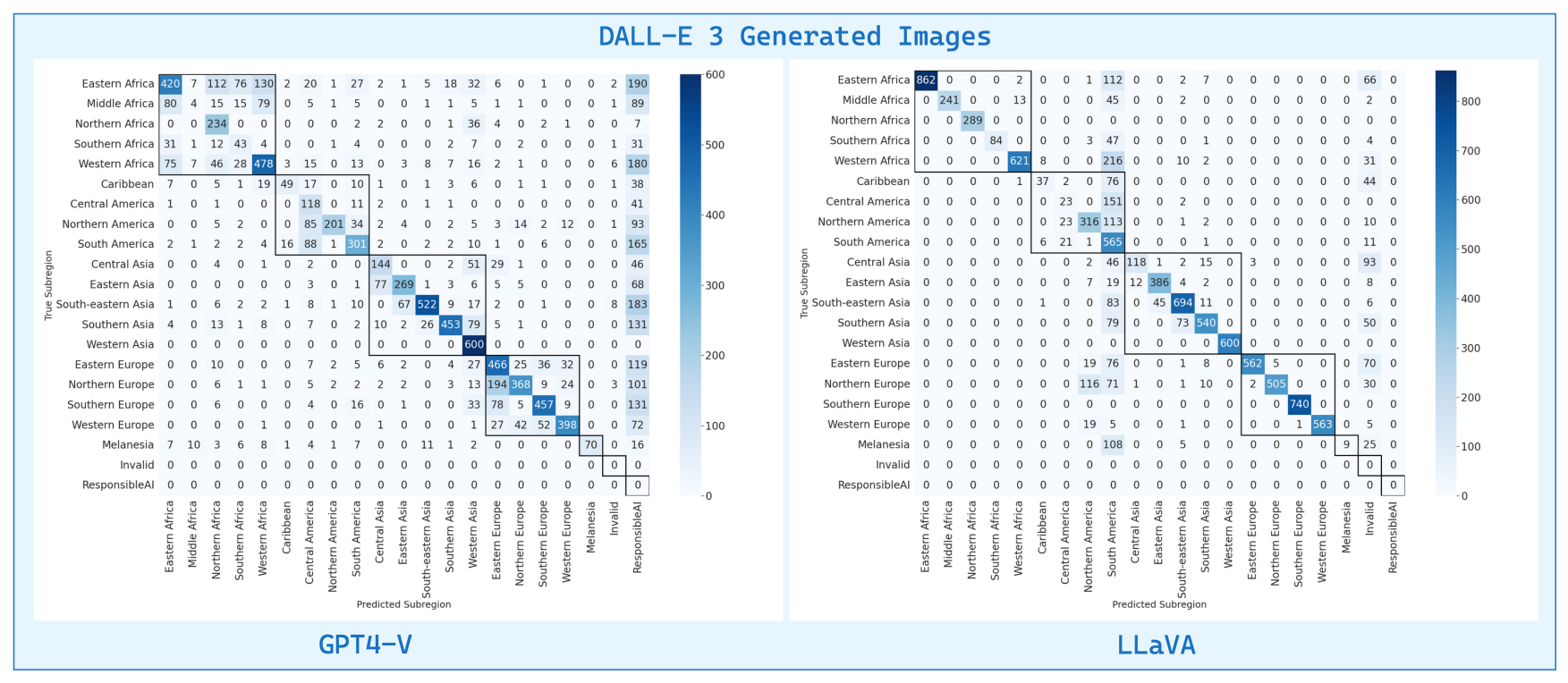}
    \caption{Confusion matrices for \gpt4v{} and \llava{} on the cultural awareness task for \dataset{} images.}
    \label{fig:conf-dalle-both}
    \vspace{-1em}
\end{figure*}

\paragraph{Confusion matrix set 2} We show confusion matrices for both \llava{} and \gpt4v{} on \dollarstreet{} images.
\begin{figure*}[t]
    \centering
    \includegraphics[width=\textwidth]{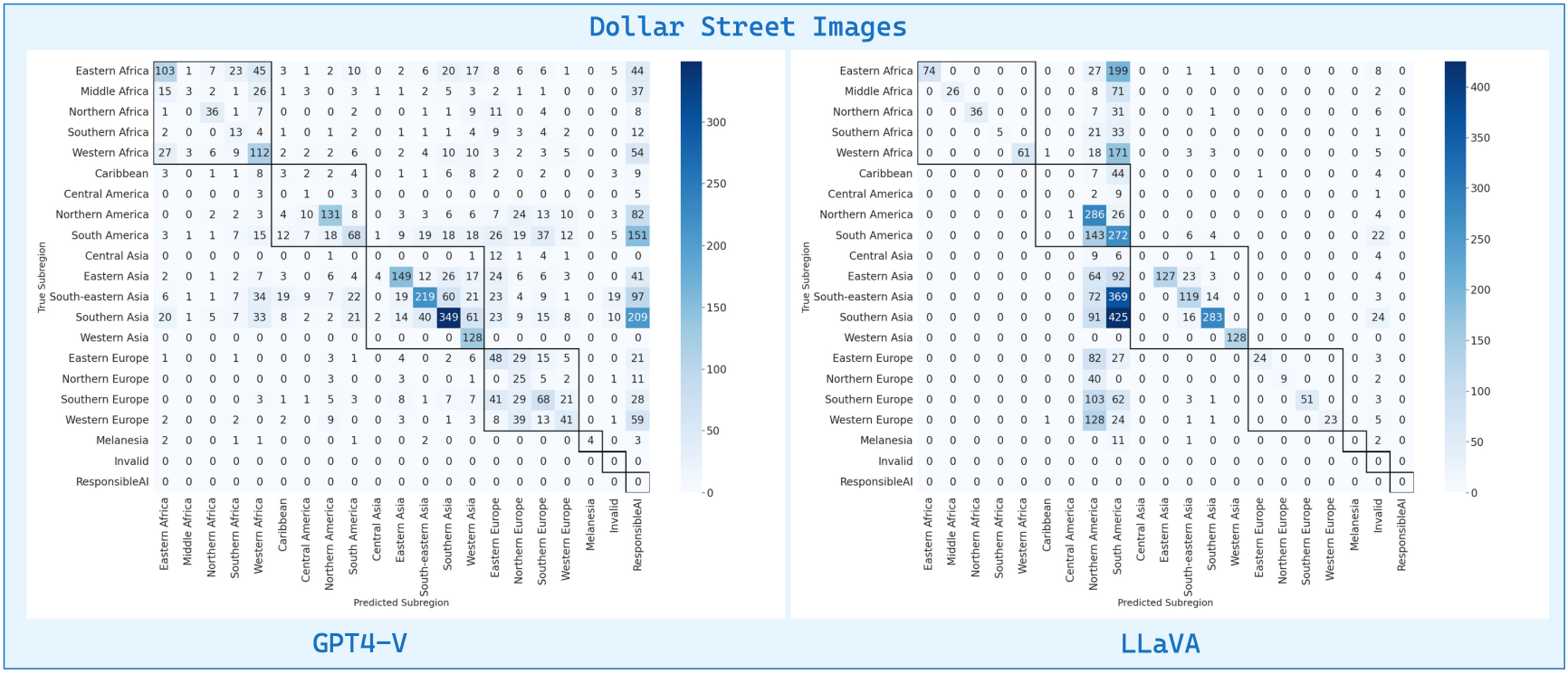}
    \caption{Confusion matrices for \gpt4v{} and \llava{} on the cultural awareness task for \dollarstreet{} images.}
    \label{fig:conf-dollar-both}
    \vspace{-1em}
\end{figure*}

\paragraph{Confusion matrix set 3} We show confusion matrices for both \llava{} and \gpt4v{} on \marvl{} images.
\begin{figure*}[t]
    \centering
    \includegraphics[width=\textwidth]{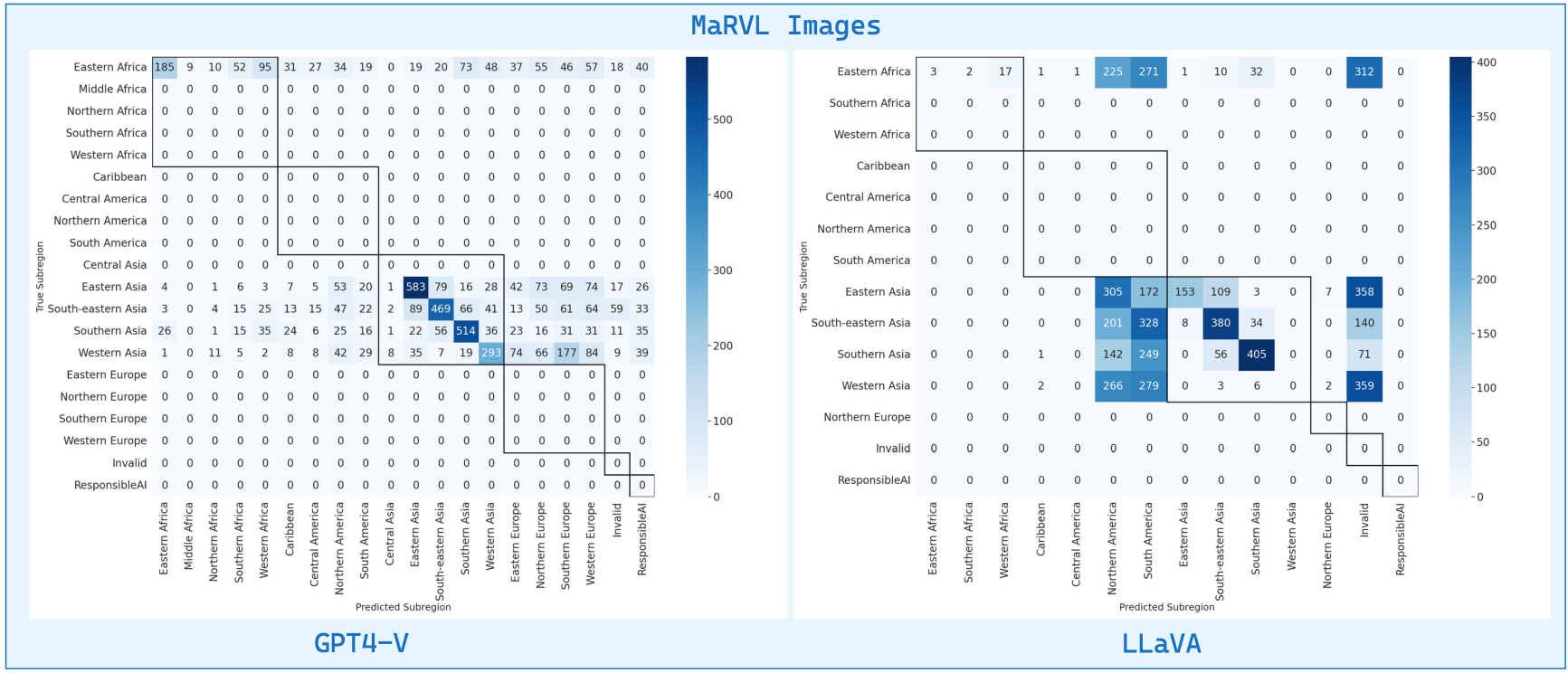}
    \caption{Confusion matrices for \gpt4v{} and \llava{} on the cultural awareness task for \marvl{} images.}
    \label{fig:conf-marvl-both}
    \vspace{-1em}
\end{figure*}

\subsubsection{Task 2 - Artifacts}
\paragraph{How many artifacts did we identify using \gpt4v{}} Table~\ref{tab:artifact-statistics} shows the counts across each of the 67 countries for the number of artifacts identified. \texttt{adj} implies unique artifacts are a combination of words and adjectives that appear before it to quantify count or color. \texttt{no\_adj} implies only the raw words identified. Figure~\ref{fig:tfidf-dist} includes a distribution of the TD-IDF scores for these (country, artifact) pairs and high scores that lie outside the range as given by the mean and the standard deviation of the distribution (i.e larger than 3.01 or smaller than 0.47) would imply strongly correlated artifacts for a given country, and there exists 4019 such items from this data. Further human filtering can be done to remove common words like table or mailbox or dresses to find unique and interesting associations like pretzels in Austria, zinnias in Bolivia and many more, some of which we report in Table~\ref{tab:artifact-interesting}.

\begin{figure*}[t]
    \centering
    \includegraphics[width=\textwidth]{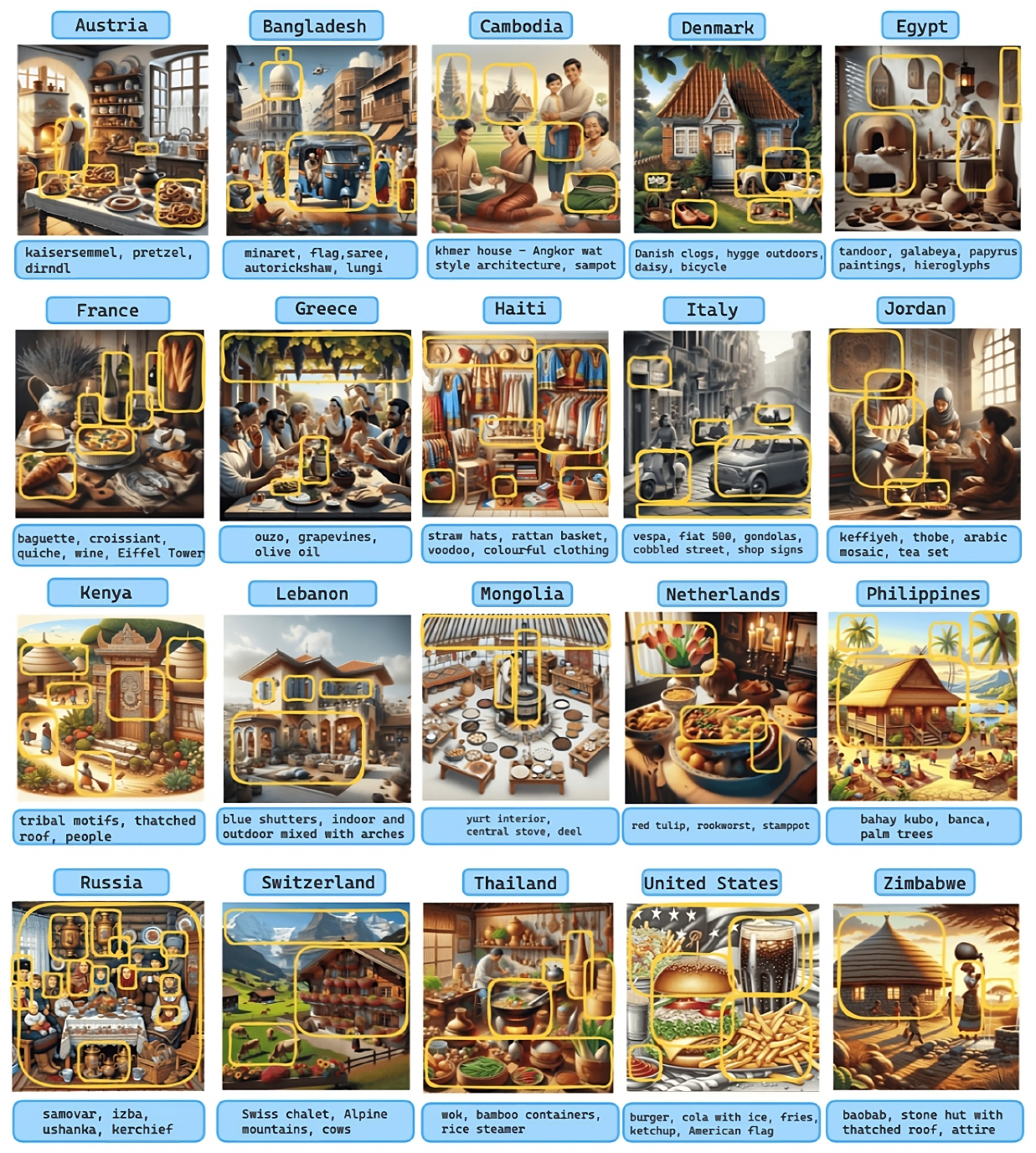}
    \caption{We identify more than 18,000 unique cultural artifacts across all countries as part of our second task, and then filter them to find salient ones. This figure shows strongest correlated artifacts for 20 randomly picked countries.\\\textcolor{orange}{Note that these associations are extracted from LMM generations and may not always be accurate.}}
    \label{fig:cultural-artifacts-all}
    \vspace{-1em}
\end{figure*}

\begin{table}[htbp]
    \scriptsize
    \centering
    \resizebox{\textwidth}{!}{
    \begin{tabular}{lcccccccccc}
        \toprule
        \textbf{} & \textbf{Austria} & \textbf{Bangladesh} & \textbf{Bolivia} & \textbf{Brazil} & \textbf{Bulgaria} & \textbf{Burkina Faso} & \textbf{Burundi} & \textbf{Cambodia} & \textbf{Cameroon} & \textbf{Canada} \\
        \midrule
        \textbf{adj} & 248 & 283 & 279 & 264 & 264 & 288 & 292 & 276 & 251 & 292 \\
        \textbf{no\_adj} & 154 & 148 & 141 & 157 & 138 & 153 & 141 & 161 & 133 & 170 \\
        \midrule
        \textbf{} & \textbf{China} & \textbf{Colombia} & \textbf{Cote d'Ivoire} & \textbf{Czech Republic} & \textbf{Denmark} & \textbf{Egypt} & \textbf{Ethiopia} & \textbf{France} & \textbf{Ghana} & \textbf{Greece} \\
        \midrule
        \textbf{adj} & 275 & 268 & 270 & 276 & 278 & 276 & 252 & 278 & 265 & 270 \\
        \textbf{no\_adj} & 124 & 134 & 146 & 160 & 168 & 152 & 139 & 157 & 137 & 151 \\
        \midrule
        \textbf{} & \textbf{Guatemala} & \textbf{Haiti} & \textbf{India} & \textbf{Indonesia} & \textbf{Iran} & \textbf{Italy} & \textbf{Jordan} & \textbf{Kazakhstan} & \textbf{Kenya} & \textbf{Kyrgyzstan} \\
        \midrule
        \textbf{adj} & 272 & 301 & 264 & 312 & 246 & 262 & 280 & 257 & 270 & 269 \\
        \textbf{no\_adj} & 163 & 160 & 147 & 172 & 123 & 135 & 158 & 139 & 149 & 147 \\
        \midrule
        \textbf{} & \textbf{Latvia} & \textbf{Lebanon} & \textbf{Liberia} & \textbf{Lithuania} & \textbf{Malawi} & \textbf{Mexico} & \textbf{Mongolia} & \textbf{Myanmar} & \textbf{Nepal} & \textbf{Netherlands} \\
        \midrule
        \textbf{adj} & 269 & 244 & 276 & 267 & 281 & 266 & 282 & 292 & 290 & 260 \\
        \textbf{no\_adj} & 156 & 133 & 163 & 159 & 152 & 133 & 156 & 155 & 156 & 146 \\
        \midrule
        \textbf{} & \textbf{Nigeria} & \textbf{Pakistan} & \textbf{Palestine} & \textbf{Papua New Guinea} & \textbf{Peru} & \textbf{Philippines} & \textbf{Romania} & \textbf{Russia} & \textbf{Rwanda} & \textbf{Serbia} \\
        \midrule
        \textbf{adj} & 279 & 254 & 270 & 279 & 265 & 276 & 272 & 244 & 288 & 250 \\
        \textbf{no\_adj} & 152 & 144 & 137 & 141 & 137 & 152 & 149 & 144 & 161 & 143 \\
        \midrule
        \textbf{} & \textbf{Somalia} & \textbf{South Africa} & \textbf{South Korea} & \textbf{Spain} & \textbf{Sri Lanka} & \textbf{Sweden} & \textbf{Switzerland} & \textbf{Tanzania} & \textbf{Thailand} & \textbf{Togo} \\
        \midrule
        \textbf{adj} & 291 & 272 & 279 & 263 & 250 & 260 & 265 & 272 & 273 & 279 \\
        \textbf{no\_adj} & 165 & 165 & 144 & 149 & 150 & 157 & 155 & 145 & 154 & 140 \\
        \midrule
        \textbf{} & \textbf{Tunisia} & \textbf{Turkey} & \textbf{Ukraine} & \textbf{United Kingdom} & \textbf{United States} & \textbf{Vietnam} & \textbf{Zimbabwe} & \textbf{Total} & & \\
        \midrule
        \textbf{adj} & 249 & 254 & 267 & 281 & 273 & 291 & 311 & 18212 & & \\
        \textbf{no\_adj} & 133 & 132 & 148 & 181 & 162 & 163 & 166 & 10035 & & \\
        \bottomrule
    \end{tabular}}
    \caption{Salient Artifact Statistics (\textbf{adj} indicates descriptors like color, etc are part of the artifact name, whereas \textbf{no\_adj} indicates the artifact name does not have such descriptors)}
    \label{tab:artifact-statistics}
\end{table}

\paragraph{Color associations on \dalle{} images} We find that countries are more likely to be associated with particular colors, with some showing prominently strong associations.
\begin{figure*}[t]
    \centering
    \includegraphics[width=\textwidth]{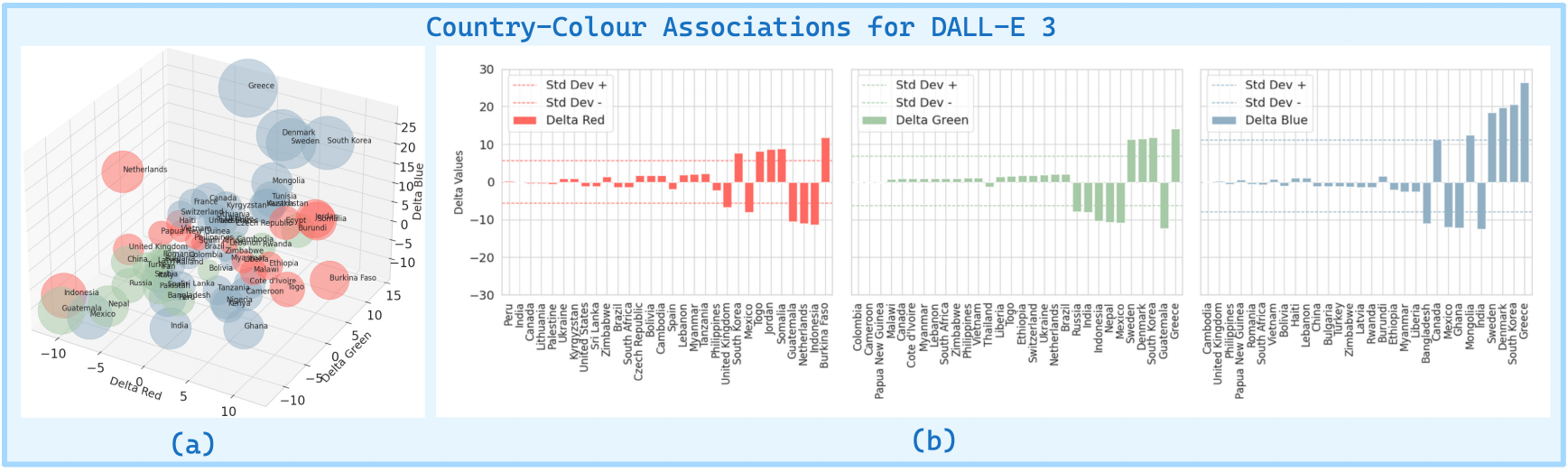}
    \caption{We explore how countries are distributed on a color spectrum by first calculating a global average RGB vector for \dataset{} images and then defining deltas along each axes aggregated at the country level. \textbf{Takeaway}: We find interesting associations - Greece is strongly correlated with blue, Burkina Faso with red.}
    \label{fig:rgb-deltas}
    \vspace{-1em}
\end{figure*}

\paragraph{People-count associations on \dalle{} images} Distributions validated by humans do not always correlate with actual population statistics.
\begin{figure*}[h]
    \centering
    \includegraphics[width=\textwidth]{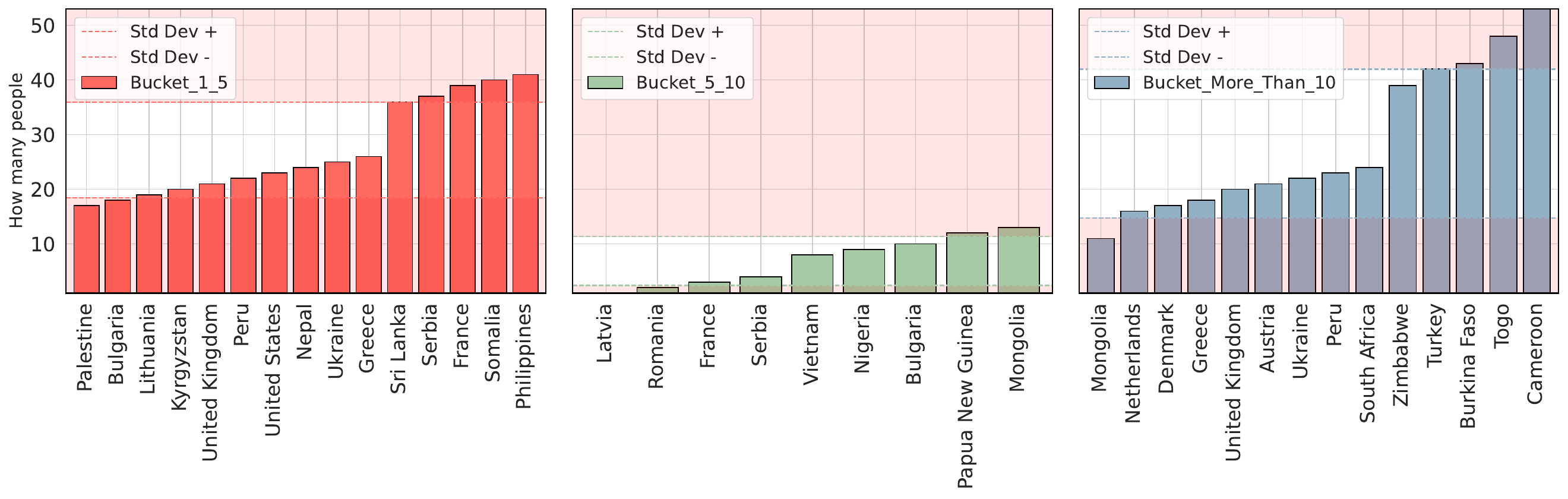}
    \caption{Here, we look at buckets of people counts in \dataset{} images aggregated at the country level, each of the subplots representing one bucket. \textbf{Takeaway}: Counts of people in images may not always accurately reflect population densities of the corresponding countries to scale.}
    \label{fig:people-counts}
    \vspace{-1em}
\end{figure*}

\paragraph{Interesting associations} We show examples of interesting associations identified by models and humans at the country level, for our artifact extraction task.
\begin{longtable}{p{3cm}p{12cm}}
    \caption{Interesting associations and their explanations for various countries.\\\textcolor{orange}{Note that these associations are extracted from LMM generations and may not always be accurate.}}
    \label{tab:artifact-interesting}\\
    \toprule
    \textbf{Country} & \textbf{Interesting Associations and Explanations} \\
    \midrule
    \endfirsthead

    \toprule
    \textbf{Country} & \textbf{Interesting Associations and Explanations} \\
    \midrule
    \endhead

    \midrule
    \multicolumn{2}{r}{\textit{Continued on next page}} \\
    \midrule
    \endfoot

    \bottomrule
    \endlastfoot

    \textbf{Austria} & 
    \textbf{Dirndl}: A traditional dress worn in Austria and parts of Germany. \\
    & \textbf{Pretzel}: A type of baked bread product, often associated with German-speaking countries. \\
    & \textbf{Lederhosen}: Traditional leather shorts worn by men in the Alpine regions. \\
    \midrule
    \textbf{Bangladesh} & 
    \textbf{Lungi}: A traditional garment worn by men, usually a wraparound skirt. \\
    & \textbf{Kurti}: A traditional garment worn by women, often paired with leggings or a skirt. \\
    & \textbf{Harmonium}: A musical instrument commonly used in South Asian music. \\
    \midrule
    \textbf{Bolivia} & 
    \textbf{Zinnias}: A type of flower native to the region, known for its bright colors and significance in local celebrations. \\
    & \textbf{Llama}: A domesticated South American camelid, significant in Bolivian culture. \\
    & \textbf{Chullos}: Knitted hats, typically with ear flaps, that are traditional to the Andes. \\
    \midrule
    \textbf{Brazil} & 
    \textbf{Bikini}: Associated with the famous beaches of Brazil. \\
    & \textbf{Lychee}: A tropical fruit found in Brazil. \\
    & \textbf{Samba}: A Brazilian music genre and dance style. \\
    \midrule
    \textbf{Bulgaria} & 
    \textbf{Spanakopita}: A savory pastry filled with spinach and feta cheese. \\
    & \textbf{Moussaka}: A layered dish with eggplant, potatoes, and minced meat. \\
    & \textbf{Terracotta}: Refers to clay-based unglazed or glazed ceramic. \\
    \midrule
    \textbf{Cameroon} & 
    \textbf{Kaftans}: A type of long robe worn in many African countries. \\
    & \textbf{Fufu}: A dough-like food made from cassava or yams. \\
    & \textbf{Savanna}: A type of ecosystem common in Cameroon, characterized by grassland with scattered trees. \\
    \midrule
    \textbf{Canada} & 
    \textbf{Poutine}: A dish consisting of fries topped with cheese curds and gravy. \\
    & \textbf{Moose}: A large mammal found in Canada. \\
    & \textbf{Snowmobile}: A vehicle designed for travel on snow, common in Canadian winters. \\
    \midrule
    \textbf{China} & 
    \textbf{Changshan}: A traditional Chinese garment for men. \\
    & \textbf{Baozi}: A type of Chinese steamed bun with fillings. \\
    & \textbf{Lion Dance}: A traditional dance in Chinese culture performed during the Lunar New Year and other cultural events. \\
    \midrule
    \textbf{Ethiopia} & 
    \textbf{Injera}: A sourdough flatbread and a staple food in Ethiopia. \\
    & \textbf{Wat}: A traditional Ethiopian stew. \\
    & \textbf{Shawl}: Often worn by Ethiopian women as part of traditional attire. \\
    \midrule
    \textbf{France} & 
    \textbf{Camembert}: A famous French cheese. \\
    & \textbf{Baguette}: A long, thin loaf of French bread. \\
    & \textbf{Beret}: A soft, round, flat-crowned hat associated with French culture. \\
    \midrule
    \textbf{Germany} & 
    \textbf{Oktoberfest}: An annual beer festival and cultural event in Munich. \\
    & \textbf{Bratwurst}: A type of German sausage. \\
    & \textbf{Dirndl}: Traditional dress worn by women during Oktoberfest and other occasions. \\
    \midrule
    \textbf{Greece} & 
    \textbf{Toga}: A garment worn in ancient Greece. \\
    & \textbf{Dolma}: A dish made of grape leaves stuffed with rice or meat. \\
    & \textbf{Moussaka}: A layered dish with eggplant, meat, and béchamel sauce. \\
    \midrule
    \textbf{India} & 
    \textbf{Sari}: A traditional garment worn by women. \\
    & \textbf{Lassi}: A yogurt-based drink. \\
    & \textbf{Rangoli}: A form of art created on the floor using colored rice, sand, or flower petals. \\
    \midrule
    \textbf{Japan} & 
    \textbf{Kimono}: A traditional Japanese garment. \\
    & \textbf{Sushi}: A popular Japanese dish. \\
    & \textbf{Tatami}: A type of mat used as a flooring material in traditional Japanese rooms. \\
    \midrule
    \textbf{Mexico} & 
    \textbf{Sombrero}: A wide-brimmed hat traditionally worn in Mexico. \\
    & \textbf{Tacos}: A traditional Mexican dish. \\
    & \textbf{Guacamole}: A Mexican avocado-based dip or spread. \\
    \midrule
    \textbf{Morocco} & 
    \textbf{Tagine}: A North African dish named after the earthenware pot in which it is cooked. \\
    & \textbf{Kaftan}: A long robe worn in Morocco. \\
    & \textbf{Mint Tea}: A popular beverage in Morocco, often served as a welcoming gesture. \\
    \midrule
    \textbf{Nepal} & 
    \textbf{Topi}: A traditional hat worn in Nepal. \\
    & \textbf{Himalayas}: The mountain range running across Nepal. \\
    & \textbf{Dal Bhat}: A traditional Nepalese dish consisting of lentils and rice. \\
    \midrule
    \textbf{Peru} & 
    \textbf{Chullo}: A traditional hat with earflaps. \\
    & \textbf{Llama}: A significant animal in Peruvian culture. \\
    & \textbf{Ponchos}: Traditional clothing made from wool. \\
    \midrule
    \textbf{Thailand} & 
    \textbf{Tuk-tuk}: A common form of transportation in Thailand. \\
    & \textbf{Pad Thai}: A popular Thai noodle dish. \\
    & \textbf{Elephant}: An animal deeply ingrained in Thai culture and symbolism. \\
    \midrule
    \textbf{Togo} & 
    \textbf{Kente Cloth}: A traditional fabric made of silk and cotton, known for its vibrant colors and patterns. \\
    & \textbf{Yam Festival}: A major cultural festival celebrating the harvest of yams. \\
    & \textbf{Agbadza Dance}: A traditional dance performed during festivals and ceremonies. \\
    \midrule
    \textbf{Tunisia} & 
    \textbf{Shisha}: A popular water pipe used for smoking flavored tobacco. \\
    & \textbf{Harissa}: A spicy chili paste that is a staple in Tunisian cuisine. \\
    & \textbf{Mosaic Art}: Intricate and colorful tile art that is significant in Tunisian culture. \\
    \midrule
    \textbf{Turkey} & 
    \textbf{Evil Eye}: A common talisman believed to protect against negative energy. \\
    & \textbf{Baklava}: A sweet pastry made of layers of filo filled with nuts and honey. \\
    & \textbf{Whirling Dervishes}: A religious dance performed by Sufi practitioners. \\
    \midrule
    \textbf{Ukraine} & 
    \textbf{Pysanky}: Traditional Ukrainian Easter eggs decorated with intricate designs. \\
    & \textbf{Borscht}: A beet soup that is a key part of Ukrainian cuisine. \\
    & \textbf{Vyshyvanka}: Traditional Ukrainian embroidered shirts. \\
    \midrule
    \textbf{United Kingdom} & 
    \textbf{Afternoon Tea}: A British tradition involving tea and a variety of snacks. \\
    & \textbf{Red Telephone Box}: Iconic public telephone booths found throughout the UK. \\
    & \textbf{Fish and Chips}: A classic British dish of battered fish and fried potatoes. \\
    \midrule
    \textbf{United States} & 
    \textbf{Route 66}: A historic highway symbolizing the American road trip. \\
    & \textbf{Thanksgiving}: A national holiday celebrating the harvest and other blessings. \\
    & \textbf{Statue of Liberty}: A symbol of freedom and democracy in the US. \\
    \midrule
    \textbf{Vietnam} & 
    \textbf{Ao Dai}: A traditional Vietnamese dress for women. \\
    & \textbf{Pho}: A Vietnamese noodle soup that is a staple dish. \\
    & \textbf{Conical Hat (Non La)}: A traditional hat made of bamboo and palm leaves. \\
    \midrule
    \textbf{Zimbabwe} & 
    \textbf{Mbira}: A traditional musical instrument also known as the thumb piano. \\
    & \textbf{Great Zimbabwe}: The ruins of an ancient city, significant in Zimbabwean history. \\
    & \textbf{Victoria Falls}: One of the largest and most famous waterfalls in the world, located on the border between Zimbabwe and Zambia. \\
\end{longtable}

\begin{longtable}{l|l}
    \caption{Cultural artifacts for various countries based on human annotations.\\\textcolor{orange}{Note that these artifacts are based on subjective perceptions of our human annotators and may not be completely accurate always.}}
    \label{tab:artifact-human}\\
    \hline
    \textbf{Country} & \textbf{Cultural Artifacts} \\
    \hline
    \endfirsthead

    \hline
    \textbf{Country} & \textbf{Cultural Artifacts} \\
    \hline
    \endhead

    \hline
    \multicolumn{2}{r}{\textit{Continued on next page}} \\
    \hline
    \endfoot

    \hline
    \endlastfoot

    \textbf{Austria} & beer, sausage, dirndl \\
    \textbf{Bangladesh} & rice, saree, fish \\
    \textbf{Bolivia} & colorful clothes, poncho, hats \\
    \textbf{Brazil} & brazilian flag, tropical fruit, colorful pottery \\
    \textbf{Bulgaria} & clothing, rugs, door \\
    \textbf{Burkina Faso} & dry, black people, straw basket \\
    \textbf{Burundi} & rice, beans, bananas \\
    \textbf{Cambodia} & buddhism, buddhist art, clothing \\
    \textbf{Cameroon} & african people, bananas, beans \\
    \textbf{Canada} & maple leaf, canadian flag, poutine \\
    \textbf{China} & characters, chinese food, lanterns \\
    \textbf{Colombia} & coffee, rice, avocado \\
    \textbf{Cote d'Ivoire} & black people, dry, african outfit \\
    \textbf{Czech Republic} & beer, dress, czech \\
    \textbf{Denmark} & danish flag, beer, windmill \\
    \textbf{Egypt} & hieroglyphs, egyptian art, islamic clothing \\
    \textbf{Ethiopia} & coffee, colors, clay pots \\
    \textbf{France} & baguette, cheese, wine \\
    \textbf{Ghana} & black people, african necklaces, clothing \\
    \textbf{Greece} & blue and white, sea, olives \\
    \textbf{Guatemala} & mayan art, tortilla, beans \\
    \textbf{Haiti} & black people, rice, beans \\
    \textbf{India} & naan, curry, sari \\
    \textbf{Indonesia} & buddhism, rice, clothing \\
    \textbf{Iran} & islamic art, kebab, persian rug \\
    \textbf{Italy} & pizza, pasta, wine \\
    \textbf{Jordan} & clothing, arabic, islamic art \\
    \textbf{Kazakhstan} & clothing, houses, islamic art \\
    \textbf{Kenya} & african people, african art, corn \\
    \textbf{Kyrgyzstan} & clothing, islamic art, rugs \\
    \textbf{Latvia} & clothing, beer, bread \\
    \textbf{Lebanon} & arabic clothing, hummus, bread \\
    \textbf{Liberia} & rice, black people, palm trees \\
    \textbf{Lithuania} & clothing, food, beer \\
    \textbf{Malawi} & hut, corn, black people \\
    \textbf{Mexico} & sombrero, tequila, tortilla \\
    \textbf{Mongolia} & yurt, dumplings, clothing \\
    \textbf{Myanmar} & buddhist art, rice, pagoda \\
    \textbf{Nepal} & buddhist elements, hindu elements, rice \\
    \textbf{Netherlands} & windmill, cheese, dutch clothing \\
    \textbf{Nigeria} & rice, yams, african clothing \\
    \textbf{Pakistan} & clothing, curry, sombrero \\
    \textbf{Palestine} & arabic art, hummus, bread \\
    \textbf{Papua New Guinea} & black people, tropical fruit, coconut \\
    \textbf{Peru} & inca clothing, machu picchu, andes mountains \\
    \textbf{Philippines} & rice, tropical vegetation, cooking \\
    \textbf{Romania} & clothing, sheep, ceramic pots \\
    \textbf{Russia} & fur hat, warm clothes, vodka \\
    \textbf{Rwanda} & african art, beans, dark-skinned people \\
    \textbf{Serbia} & clothing, beer, sausages \\
    \textbf{Somalia} & islamic art, banana, rice \\
    \textbf{South Africa} & african art, corn, hat \\
    \textbf{South Korea} & korean characters, kimchi, korean dress \\
    \textbf{Spain} & flamenco, paella, bull fighting \\
    \textbf{Sri Lanka} & buddhist art, buddhist symbols, spicy food \\
    \textbf{Sweden} & northern european clothing, fish, snowy landscape \\
    \textbf{Switzerland} & alps, swiss cheese, chocolate \\
    \textbf{Tanzania} & african art, rice, meat \\
    \textbf{Thailand} & buddhist art, thai food, clothing \\
    \textbf{Togo} & african clothing, cloth patterns, wood carvings \\
    \textbf{Tunisia} & arabic art, couscous, arched doorways \\
    \textbf{Turkey} & turkish coffee, rugs, kebabs \\
    \textbf{Ukraine} & clothing patterns, flower designs, ukrainian food \\
    \textbf{United Kingdom} & pubs, fish and chips, tea \\
    \textbf{United States} & american flag, burgers, jeans \\
    \textbf{Vietnam} & conical hats, pho, pagodas \\
    \textbf{Zimbabwe} & thatched huts, african clothing, animal carvings \\
\end{longtable}

\subsubsection{Task 3 - Edits}
\begin{figure*}[t]
    \centering
    \includegraphics[width=\textwidth]{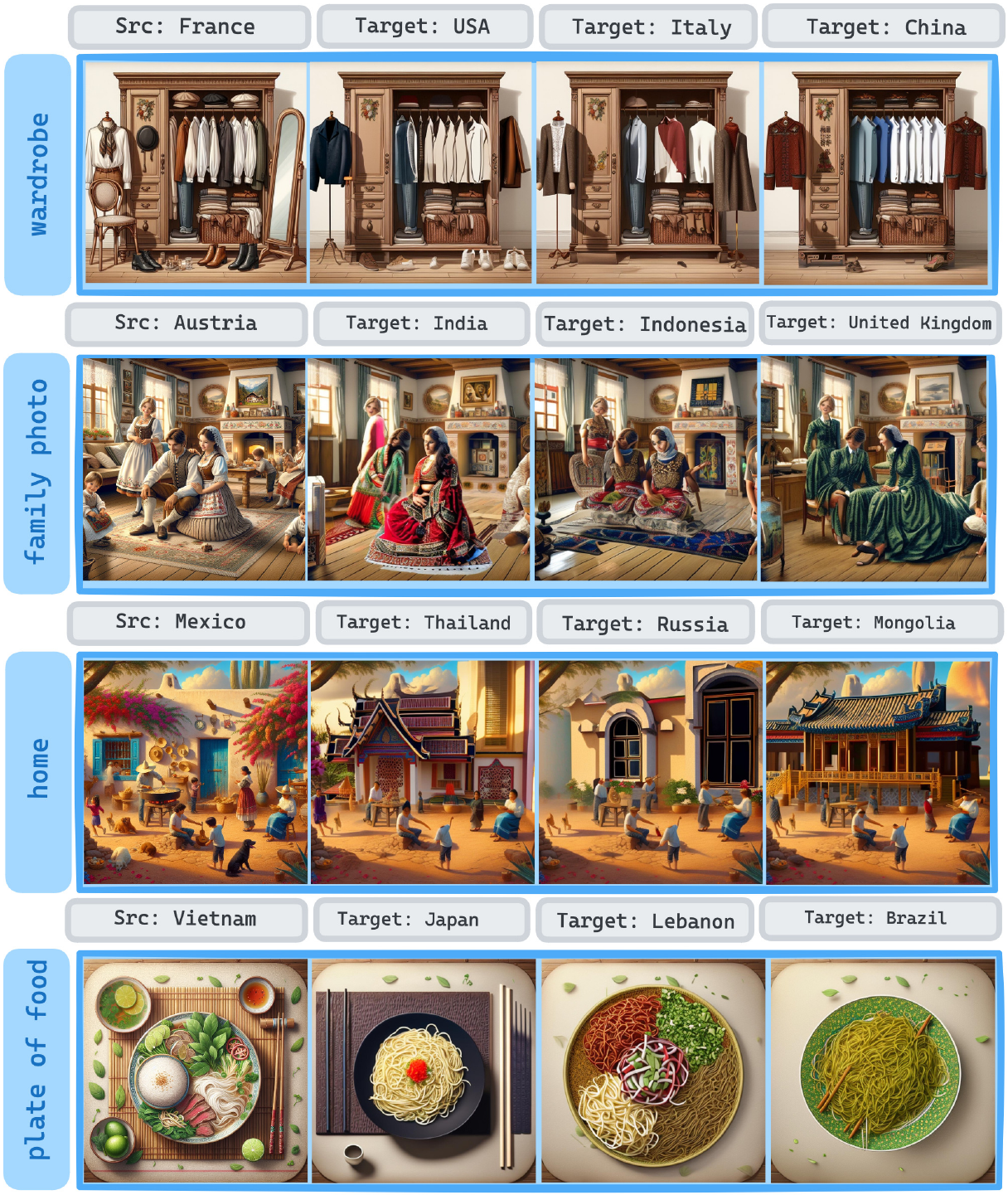}
    \caption{We show examples of edits made using our \cultureadapt{} pipeline across 4 different concept classes and 12 pairs of unique source, target combinations to illustrate both cases where our pipeline excels and also where it is limited by the parts it is composed of. For all of these edits, our metric success criteria of $\Delta_1 < 0$ and $\Delta_2 > 0$ is satisfied.}
    \label{fig:culture-adapt-edits}
    \vspace{-1em}
\end{figure*}

\paragraph{More examples of edits using \cultureadapt{}} We include more examples of edits across different concept classes and source-target pairs using our \cultureadapt{} pipeline in Figure \ref{fig:culture-adapt-edits}. As can be seen, the pipeline is only constrained by the two bottlenecks of object detection and diffusion based inpainting, which sometimes may detect objects incorrectly or not generate consistent images of human faces for example.

\paragraph{\cultureadapt{} quantitative metrics}
\begin{table}[ht]
\setlength{\tabcolsep}{1pt}
\scriptsize
\centering
\begin{tabular}{@{}lcccccc@{}}
\toprule
\textbf{Source-Target Pair}       & \multicolumn{2}{c}{\textbf{Similarity \texttt{SSIM}}}    & \multicolumn{2}{c}{\textbf{Metric \( M_1 \)}} & \multicolumn{2}{c}{\textbf{Metric \( M_2 \)}} \\ \cmidrule(lr){2-3} \cmidrule(lr){4-5} \cmidrule(lr){6-7}
                                  & \textbf{cap-edit} & \textbf{\cultureadapt{}} & \textbf{cap-edit} & \textbf{\cultureadapt{}} & \textbf{cap-edit} & \textbf{\cultureadapt{}} \\ \midrule
Brazil-India                      & 0.96 & 0.93 & 0.72 & 0.68 & 0.95 & 0.95 \\
Brazil-Nigeria                    & 0.96 & 0.93 & 0.62 & 0.47 & 0.89 & 0.91 \\
Brazil-Turkey                     & 0.96 & 0.93 & 0.69 & 0.52 & 0.95 & 0.94 \\
Brazil-USA                        & 0.96 & 0.93 & 0.28 & 0.35 & 0.91 & 0.83 \\ \cmidrule(lr){1-7}
\textit{Average}                  & \textit{0.96} & \textit{0.93} & \textit{0.58} & \textit{0.51} & \textit{0.93} & \textit{0.91} \\ \midrule
India-Brazil                      & 0.97 & 0.94 & 0.63 & 0.60 & 0.96 & 0.90 \\
India-Nigeria                     & 0.96 & 0.94 & 0.69 & 0.68 & 0.94 & 0.90 \\
India-Turkey                      & 0.97 & 0.94 & 0.56 & 0.57 & 0.92 & 0.89 \\
India-USA                         & 0.96 & 0.94 & 0.67 & 0.63 & 0.93 & 0.88 \\ \cmidrule(lr){1-7}
\textit{Average}                  & \textit{0.97} & \textit{0.94} & \textit{0.63} & \textit{0.62} & \textit{0.94} & \textit{0.89} \\ \midrule
Nigeria-Brazil                    & 0.96 & 0.92 & 0.29 & 0.41 & 0.76 & 0.85 \\
Nigeria-India                     & 0.96 & 0.92 & 0.62 & 0.67 & 0.87 & 0.93 \\
Nigeria-Turkey                    & 0.96 & 0.91 & 0.66 & 0.63 & 0.91 & 0.93 \\
Nigeria-USA                       & 0.96 & 0.92 & 0.50 & 0.60 & 0.90 & 0.91 \\ \cmidrule(lr){1-7}
\textit{Average}                  & \textit{0.96} & \textit{0.92} & \textit{0.51} & \textit{0.58} & \textit{0.86} & \textit{0.90} \\ \midrule
Turkey-Brazil                     & 0.97 & 0.94 & 0.46 & 0.41 & 0.89 & 0.84 \\
Turkey-India                      & 0.97 & 0.95 & 0.62 & 0.59 & 0.88 & 0.88 \\
Turkey-Nigeria                    & 0.97 & 0.94 & 0.67 & 0.64 & 0.93 & 0.90 \\
Turkey-USA                        & 0.96 & 0.94 & 0.29 & 0.43 & 0.88 & 0.89 \\ \cmidrule(lr){1-7}
\textit{Average}                  & \textit{0.97} & \textit{0.94} & \textit{0.51} & \textit{0.51} & \textit{0.89} & \textit{0.88} \\ \midrule
USA-Brazil                        & 0.97 & 0.94 & 0.53 & 0.46 & 0.88 & 0.89 \\
USA-India                         & 0.98 & 0.94 & 0.18 & 0.62 & 0.58 & 0.91 \\
USA-Nigeria                       & 0.98 & 0.94 & 0.20 & 0.46 & 0.61 & 0.84 \\
USA-Turkey                        & 0.98 & 0.94 & 0.21 & 0.46 & 0.64 & 0.87 \\ \cmidrule(lr){1-7}
\textit{Average}                  & \textit{0.97} & \textit{0.94} & \textit{0.28} & \textit{0.50} & \textit{0.68} & \textit{0.88} \\ \midrule
\textbf{Overall Average}          & \textbf{0.97} & \textbf{0.94} & \textbf{0.50} & \textbf{0.54} & \textbf{0.85} & \textbf{0.89} \\ \bottomrule
\end{tabular}
\caption{Mean Similarity (\texttt{SSIM}) Scores, Metric \( M_1 \), and Metric \( M_2 \) for \texttt{cap-edit} and \cultureadapt{} grouped by Source and Target Country}
\label{tab:combined_similarity_comparison}
\end{table}

\end{document}